\documentclass[sigplan,10pt]{acmart}
\acmSubmissionID{449}

\usepackage[]{hyperref}

\usepackage{graphicx}
\usepackage{multirow}
\usepackage{multicol}
\usepackage{enumitem}
\usepackage{soul}

\usepackage{algorithm}
\usepackage{algorithmic}

\usepackage{array}
\usepackage{tabularray}
\usepackage{graphicx}
\usepackage{multirow}
\usepackage{float}
\usepackage{vcell}

\usepackage{tikz}
\usepackage{amsmath}
\usepackage{filecontents}

\acmYear{2026}\copyrightyear{2026}
\setcopyright{cc}
\setcctype[4.0]{by}
\acmConference[EUROSYS '26]{European Conference on Computer Systems}{April 27--30, 2026}{Edinburgh, Scotland Uk}
\acmBooktitle{European Conference on Computer Systems (EUROSYS '26), April 27--30, 2026, Edinburgh, Scotland Uk}
\acmDOI{10.1145/3767295.3769351}
\acmISBN{979-8-4007-2212-7/26/04}



\begin{document}

\date{}

\title[FlexiQ: Adaptive Mixed-Precision Quantization for Latency/Accuracy Trade-Offs]{FlexiQ: Adaptive Mixed-Precision Quantization for Latency/Accuracy Trade-Offs in Deep Neural Networks} 
\author{Jaemin Kim}
\affiliation{
  \institution{\small Seoul National University}
  \country{South Korea}
}
\email{woals174@snu.ac.kr}

\author{Hongjun Um}
\affiliation{
  \institution{\small Hanyang University}
  \country{South Korea}
}
\email{hongjunum@hanyang.ac.kr}

\author{Sungkyun Kim}
\affiliation{
  \institution{\small Hanyang University}
  \country{South Korea}
}
\email{cheezestick@hanyang.ac.kr}

\author{Yongjun Park}
\affiliation{
  \institution{\small Yonsei University}
  \country{South Korea}
}
\email{yongjunpark@yonsei.ac.kr}

\author{Jiwon Seo}
\authornote{Corresponding author}
\affiliation{
  \institution{\small Seoul National University}
  \country{South Korea}
}
\email{seojiwon@snu.ac.kr}

\begin{abstract}
Neural networks commonly execute on hardware accelerators such as NPUs and GPUs for their size and computation overhead. These accelerators are costly and it is hard to scale their resources to handle real-time workload fluctuations.

We present FlexiQ, an adaptive mixed-precision quantization scheme for computer vision models. FlexiQ selectively applies low-bitwidth computation to feature channels with small value ranges and employs an efficient bit-lowering method to minimize quantization errors while maintaining inference accuracy. Furthermore, FlexiQ adjusts its low-bitwidth channel ratio in real time, enabling quantized models to effectively manage fluctuating inference workload.

We implemented FlexiQ prototype, including the mixed-precision inference runtime on our custom NPU and GPUs. Evaluated on eleven convolution- and transformer-based vision models, FlexiQ achieves on average 6.6\% higher accuracy for 4-bit models with finetuning and outperforms four state-of-the-art quantization techniques. Moreover, our mixed-precision models achieved an efficient accuracy-latency trade-off, with the 50\% 4-bit model incurring only 0.6\% accuracy loss while achieving 40\% of the speedup of the 100\% 4-bit model over 8-bit model.
Latency evaluations on our NPU and GPUs confirmed that FlexiQ introduces minimal runtime overhead, demonstrating its hardware efficiency and overall performance benefits.
\end{abstract}

\maketitle

\section{Introduction}

Recently, neural networks have grown significantly in size and complexity, leading to high computational overhead and inference latency. As a result, these networks are typically served on accelerators such as NPUs and GPUs. However, these accelerators are costly, and scaling their resources to handle real-time workload fluctuations -- especially under high, bursty workloads -- is challenging\,\cite{li2023alpaserve, romero2021infaas, 234998, oh2024exegpt}.

On the other hand, compression techniques such as pruning\,\cite{conf/eccv/HuangW18, conf/iclr/FrankleC19}, weight sharing\,\cite{han2015deep, chen2015compressing}, and quantization\,\cite{DBLP:conf/cvpr/WuLWHC16, pmlr-v119-wang20c, lin2024awq, kim2020robust} also aim to accelerate neural network inference. 
Particularly, quantization is widely adopted for its broad applicability and hardware efficiency. One drawback, however, is the loss of accuracy, especially under extremely low-bitwidth, such as 4-bits or fewer\,\cite{jin2024comprehensive,nagel2021white,lin2023bit}. To address this issue, mixed-precision quantization methods have been proposed, assigning different bitwidth to each layer\,\cite{dong2019hawq, dong2020hawq, habi2020hmq, hu2021opq, qu2020adaptive,wang2019haq,yao2021hawq, li2021brecq, liu2021post, yang2021fracbits, chauhan2023post, cai2020zeroq} base on, for example, a layer's computational sensitivity\,\cite{dong2019hawq}. Yet, fully computing a layer with 4-bit quantization can differ significantly from higher-bitwidth computations, potentially impacting overall accuracy.

We aim to tackle both challenges: the accuracy loss from low-bit quantization and the difficulty of adapting to dynamic workloads. To this end, we propose FlexiQ, an efficient quantization scheme for computer vision models. FlexiQ introduces a fine-grained, mixed-bit computation at the feature-channel level, applying low-bitwidth computation to selected channels and high-bitwidth to the rest. Moreover, it supports runtime adjustment of the low-bitwidth computation ratio, enabling real-time latency tuning to accommodate fluctuating inference workloads. By carefully selecting which channels use low-bitwidth computation and applying a customized bit extraction method, FlexiQ delivers an efficient and dynamic trade-off between latency and accuracy.

More specifically, FlexiQ exploits the diverse value ranges of feature channels in computer vision models. By selectively applying low-bitwidth quantization to channels with small value ranges, it reduces the quantization error introduced by low-bitwidth computation. 
Furthermore, we group multiple feature channels and apply the same quantization setting to optimize hardware efficiency while enabling runtime bitwidth adjustments for accuracy–latency trade-offs.

We evaluate FlexiQ across eleven computer vision models, comparing its performance against four quantization schemes. The results show that FlexiQ outperforms these schemes while offering an efficient real-time accuracy-latency trade-off. The contributions of this paper are as follows.

\noindent
{\bf Efficient quantization with real-time bitwidth control.} 
We propose FlexiQ, a mixed-precision quantization scheme that enables runtime bitwidth adjustment. By varying the ratio of low-bitwidth computation in real-time, FlexiQ dynamically optimizes latency to accommodate fluctuating workloads, with only a slight accuracy trade-off. As it targets feature channels with small value ranges and employs an efficient bit-lowering method, FlexiQ keeps accuracy loss to a minimum even as the ratio of low-bitwidth computation increases.

\noindent
{\bf Implementation on NPUs and GPUs.}
We implemented our mixed-precision runtime with real-time bitwidth adjustment for NPUs and GPUs. 
For NPUs, we extended DNNWeaver v2\,\cite{sharma2016dnnweaver} to support mixed-precision calculations with runtime precision switching.
On GPUs, we built a mixed-precision GEMM kernel based on Nvidia’s CUTLASS\,\cite{thakkar_cutlass_2023}. Employing a well-known nested pipeline architecture, we overlap cache and register loading with multiply-add operations. 
The kernel maps channels with the same bitwidth to a single warp, which extracts {\it effective} bits, uses the MMA PTX instruction for multiply-add, and applies bit-shifted accumulation. These implementations show that FlexiQ can be integrated with light-weight hardware and software overhead.

\noindent
{\bf Extensive evaluation.} We evaluated FlexiQ using five convolutional neural networks (CNNs) and six vision transformers, comparing it to four state-of-the-art quantization schemes. FlexiQ achieved on average 6.6\% percentage point accuracy improvement over the next best method using 4-bit quantization. 
Our mixed-precision models also provide an efficient accuracy-latency trade-off: a 50\% 4-bit / 50\% 8-bit model incurs only a 0.6\% average accuracy loss compared to the full-precision baseline, while achieving 40\% of the speedup gained by the 100\% 4-bit model over the 8-bit model. We further evaluated latency on our custom NPU and GPU using our optimized mixed-precision GeMM kernel, confirming the performance benefits of mixed-precision inference. Under real-world inference traces with fluctuating request rates, FlexiQ achieved near 8-bit accuracy (within 0.1\% on average) and median latency comparable to 4-bit inference.

The rest of the paper is organized as follows. Section\,\ref{sec:background} presents the background and motivation. Section\,\ref{sec:overview} provides an overview of FlexiQ. Section\,\ref{sec:ptq} describes the quantization method used in FlexiQ, while Section\,\ref{sec:postprocessing} covers memory optimization in post-processing. Section\,\ref{sec:qat} explains our finetuning approach with a specialized loss function. Section\,\ref{sec:npu-gpu-impl} details our NPU and GPU implementation. Section\,\ref{sec:eval} presents the evaluation results, and Section\,\ref{sec:conclusion} concludes the paper.

\section{Background and Motivation}
\label{sec:background}

\subsection{Overview of Quantization}
\label{sec:Quant}

Quantization lowers the precision of inference computations from 32-bit floating point to 8-bit or lower to reduce inference latency. The most common and straightforward method is uniform quantization, which maps a floating-point value $\textbf{x}$ to a quantized value  $\textbf{x}_q$ as follows:
\setlength{\abovedisplayskip}{4pt}
\setlength{\belowdisplayskip}{4pt}
\begin{gather}
\textbf{x}_q = \text{clip}\left(\left\lfloor\frac{\textbf{x}}{S}\right\rceil, Q_n, Q_p\right) \label{eq:1}
\end{gather}

\noindent
Here, $S$ is a scaling factor (i.e., the size of one quantization step), and $[Q_n, Q_p]$ is the integer range supported by the target bitwidth (e.g., $[-128, 127]$ for 8-bit quantization).

Uniform quantization introduces minimal runtime overhead and supports efficient integer-only inference\,\cite{jacob2018quantization}. However, its accuracy degrades largely at very low bitwidths (e.g., 4-bit), due to larger quantization steps ($S$) that increase the error between the original and quantized values ($|x-x_q|$). Moreover, it lacks the flexibility to adapt inference latency in response to fluctuating request rates -- an important requirement in real-world serving systems. %

Mixed-precision quantization combines low- and high-bitwidth quantization to mitigate the accuracy loss observed in uniform quantization. Early approaches, such as HAWQ\,\cite{dong2019hawq, dong2020hawq, yao2021hawq} and HAQ\,\cite{wang2019haq}, adopted a layer-wise strategy by finding precision-insensitive layers and applying low-bitwidth quantization only to those layers. While this method allows mixing different precisions, computing an entire layer at very low bitwidth (i.e., 4-bit) still leads to noticeable accuracy degradation, as we show in Section\,\ref{sec:motiv}.

More recent methods, including Atom\,\cite{zhao2024atom} and LLM.int8() \cite{dettmers2022gpt3}, extend mixed-precision quantization to operate within individual layers. This finer-grained approach makes it harder to support integer-only execution. For instance, matrix multiplications (e.g., GeMM) can be performed entirely with integer arithmetic only, if all values are quantized using the same scale factor. When values are quantized with different scales or bitwidths, the scale factor cannot be factored out of the summation, as shown below:
\[
y = \sum_i (S_{1,i} q_{1,i})(S_{2,i} q_{2,i}) \ne S_1 S_2 \sum_i q_{1,i} q_{2,i}
\]
\noindent
To address this, these methods apply high-bitwidth computations only to a small number of outliers. However, since they are not designed with latency-accuracy trade-offs in mind, they do not support dynamically adjusting the low/high bitwidth ratio in response to changing request rates.

\subsection{Related Works}
\label{sec:related}

To improve accuracy under extreme low-bit quantization, mixed-precision methods have been introduced\,\cite{dong2019hawq, wang2019haq, dong2020hawq, yao2021hawq}. 
As discussed earlier, they typically assingn low- and high-bitwidth to different layers to preserve accuracy.
While effective, these methods produce a single static model, which limits adaptability to workload fluctuations. Supporting such adaptation requires managing multiple model instances, incurring substantial storage and runtime overhead.

More recently, multi-precision quantization has been proposed to enable inference at multiple precisions from a single model. RobustQuant\,\cite{chmiel2020robust} and AnyPrecision\,\cite{yu2021any} apply finetuning with the objective of preserving high accuracy under low- to high-bitwidth quantization as well as full-precision inference. PTMQ\,\cite{xu2024ptmq} instead maintains multiple sets of scale factors within one model to support low- to high-bitwidth inference. These approaches, however, still require storing the full-precision model and dynamically quantizing it at runtime, which introduces memory overhead and nontrivial performance costs when switching precisions.

Orthogonally, runtime systems have been developed to balance latency, throughput, and accuracy under workload fluctuations\,\cite{ghodrati2020planaria, li2023alpaserve, stojkovic2025dynamollm}. MArk\,\cite{234998}, for example, dynamically adjusts batch sizes and resource allocation to trade-off throughput (cost) against latency, while INFaaS\,\cite{romero2021infaas} maintains multiple models with varying accuracies and execution times, selecting among them based on request constraints. 
In the context of LLM inference, DistServe\,\cite{zhong2024distserve} and ExeGPT\,\cite{oh2024exegpt} trade off per-request decoding latency with overall throughput using scheduling optimizations, similar to those applied in training\,\cite{shoeybi2019megatron, oh2022out}.

\subsection{Observation and Preliminary Analysis}
\label{sec:motiv}
Recent techniques exploiting domain information have shown to improve quantization accuracy. For Large Language Models (LLMs), a small subset of token embedding dimensions have significantly larger values than others, suggesting that shifting scale variance from activations to weights can improve accuracy\,\cite{XiaoLSWDH23}. In computer vision models, channel-wise quantization addresses wide variation in quantization ranges across weight parameters of different output channels. By assigning different scale values to each output channel, quantization errors are effectively reduced\,\cite{gholami2022survey}.

\begin{figure}[t]
\centering
\includegraphics[width=0.90\linewidth]{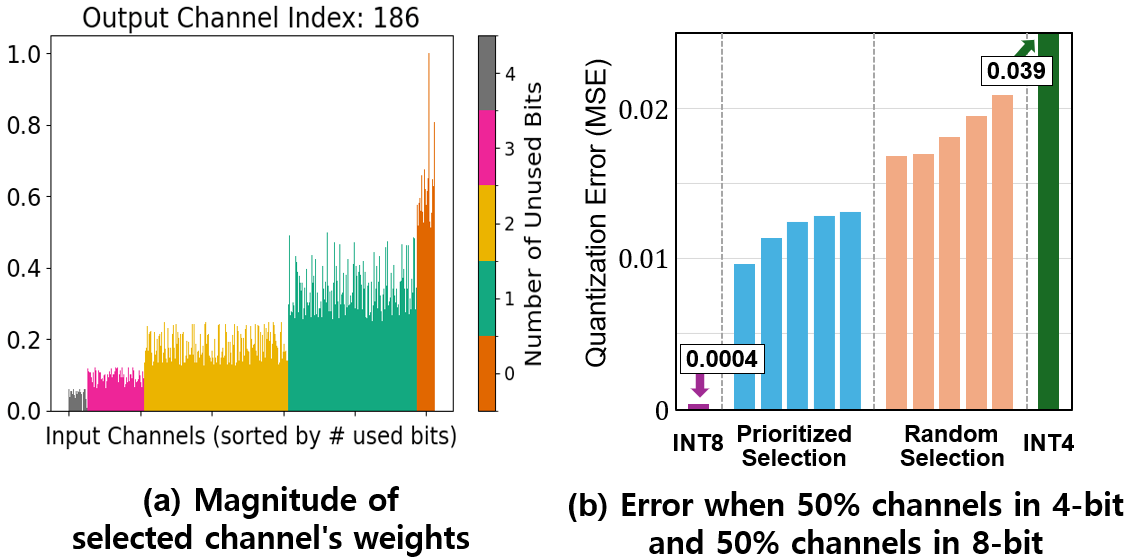}
\caption{Number of unused bits in weight parameters (ResNet-50 layer 51) under 8-bit quantization due to small value ranges (left). Quantization error for this layer when applying 50\% 4-bit quantization with/without exploiting these unused bits (right, bit extraction in Section\,\ref{sec:bit-lowering} used).}
\label{fig:FeatureChannel}
\end{figure}

In these vision models, we observed that quantization ranges still vary largely across parameters within the same output channels. Figure\,\ref{fig:FeatureChannel}(a) shows this trend in a randomly selected output channel with its parameters grouped by corresponding input (i.e., feature) channels. Many of these groups have small value ranges, and since they are quantized with the same scale value, the high bits in their quantized representation remain unused. 

To make use of the unused bits in channels with small value ranges, we applied mixed-precision quantization: 4-bit for parameter groups with small ranges (50\% of all, grouped by feature channels) and 8-bit for the rest. As shown in Figure\,\ref{fig:FeatureChannel}(b), this approach keeps the layer's computation close to that of 32-bit floating-point. We also observed that increasing the proportion of 4-bit parameters (in the feature dimension) reduces inference latency with only a modest drop in accuracy, enabling a practical trade-off between speed and precision for inference serving.

\begin{figure*}[t]
\centering
\includegraphics[width=0.90\textwidth]{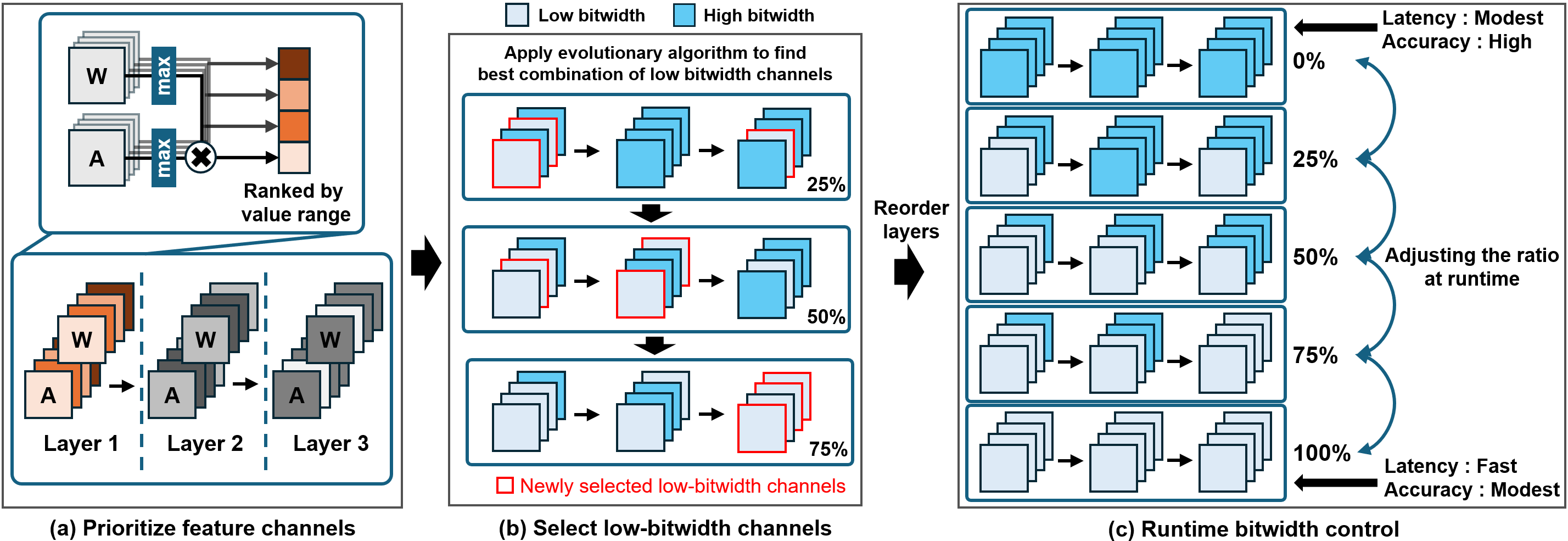}
\caption{Overview of FlexiQ: mixed-precision quantization with channel prioritization, selection, and runtime control.}
\label{fig:Overview}
\end{figure*}

\vspace{1mm}
\noindent
{\bf Problem Statement.} Building on these observations and analysis, we define the focus of our study as optimizing the trade-off between accuracy and latency by selectively applying low-bitwidth quantization to a subset of $\textit{feature channels}$. Specifically, we begin with a high-bitwidth quantized model and identify a subset of features to compute using low-bitwidth quantization, targeting a specific percentage of the total calculations. To support real-time transitions between high-bitwidth and low-bitwidth quantization, we ensure that the bits used for low-bitwidth quantization are a strict subset of those used in the high-bitwidth model.

More formally, our optimization problem of finding a subset of weight parameters for low-bitwidth computation is formulated as below (assuming a single-layer model):
\vspace{1mm}
$$
\text{argmin}_{\beta,W} \left\| Y - \sum_{i=1}^{C} \left[ (1-\beta_i) X^{\text{high}}_i W^{\text{high}}_i + \beta_i X^{\text{low}}_i W^{\text{low}}_i \right] \right\|_F^2 \\
$$
$$\text{subject to } \beta \in \{0, 1\}^C, \quad \|\beta\|_0 \geq \text{target ratio} 
$$
\noindent
where $F$ is the Frobenius norm, $C$ is the number of channels in the layer, $W^{\text{high}}$ and $W^{\text{low}}$ are high- and low-bitwidth versions of weight parameters (we require that obtaining $W^{\text{low}}$ from $W^{\text{high}}$ is inexpensive), and $\beta$ is a vector of flags for the channels, indicating low-bitwidth computation.

\section{Overview of FlexiQ}
\label{sec:overview}

The problem formulated in Section\,\ref{sec:motiv} is a combinatorial optimization problem and is NP-hard. Thus we designed a heuristic method for selecting feature channels and applying low-bitwidth quantization efficiently. Our proposed method, called FlexiQ, allows real-time adjustments to the ratio of low-bitwidth computations, thus supporting a dynamic trade-off between latency and accuracy.

The core idea of FlexiQ is to find the feature channels where the activation and corresponding weight parameters have small value ranges for most of input data. We selectively apply low-bitwidth quantization to these channels to take advantage of the unused bits in their high-bitwidth representations. By excluding these unused bits when lowering the bitwidth, we increase effective precision and reduce quantization errors. To avoid error amplification across layers, we use an evolutionary algorithm that considers inter-layer interactions when selecting channels for low-bitwidth quantization. Furthermore, we make it possible to dynamically detect channel value ranges and adjust the bit extraction strategy to further reduce quantization error.

An overview of how FlexiQ works is shown in Figure\,\ref{fig:Overview}. Starting with a high-bitwidth quantized model, we rank all feature channels by their estimated quantization error based on value ranges. These ranges also guide the bit-lowering method applied to each channel. 
Using the estimated errors, our evolutionary algorithm selects progressively larger sets of channels for low-bitwidth quantization while accounting for inter-layer error amplification. The resulting models, each with a different ratio of low-bitwidth channels, enable flexible latency-accuracy trade-offs during inference serving to adapt to varying request rates.

To improve the accuracy of predominantly low-bitwidth quantized models, FlexiQ supports finetuning. We design a specialized loss function that preserves the accuracy of the fully high-bitwidth model while improving the performance of fully or predominantly low-bitwidth models. After finetuning, we apply the same evolutionary algorithm for channel selection. As detailed in Section\,\ref{sec:ptq}, the algorithm incorporates domain knowledge to efficiently explore the selection space.

At runtime, FlexiQ supports adjusting the low-bitwidth ratio efficiently, since low-bitwidth values are a strict subset of high-bitwidth representations and can be extracted from them. To further optimize for hardware efficiency, we group multiple channels and assign the same bitwidth to each group (details in Section\,\ref{sec:npu-gpu-impl}). Together, these features make FlexiQ practical for adaptive inference in real-world serving scenarios.

\section{Channel Selection and Bit-Lowering}
\label{sec:ptq}

Vision models have diverse value ranges across their feature channels (shown in Section\,\ref{sec:motiv} and further discussed in Section\,\ref{sec:eval}), and FlexiQ exploits this diversity to apply low-bitwidth quantization effectively. This requires 1) selecting channels for low-bitwidth computation while accounting for error amplification due to inter-layer interactions, and 2) optimizing the bit-lowering to minimize quantization errors. We first describe the bit-lowering optimization, followed by the channel selection process.

\subsection{Effective Bit Extraction (Bit-Lowering)}
\label{sec:bit-lowering}
We begin by presenting our bit-lowering method, which helps improve the effectiveness of the channel selection algorithm. For feature channels with small value ranges, we find the unused bits of their values in advance and quantize them using the most significant used bits. This approach increases the {\it effective} bitwidth and thus reduces quantization error when applying low-bitwidth quantization.

\begin{figure}[t]
\centering
\includegraphics[width=0.95\linewidth]{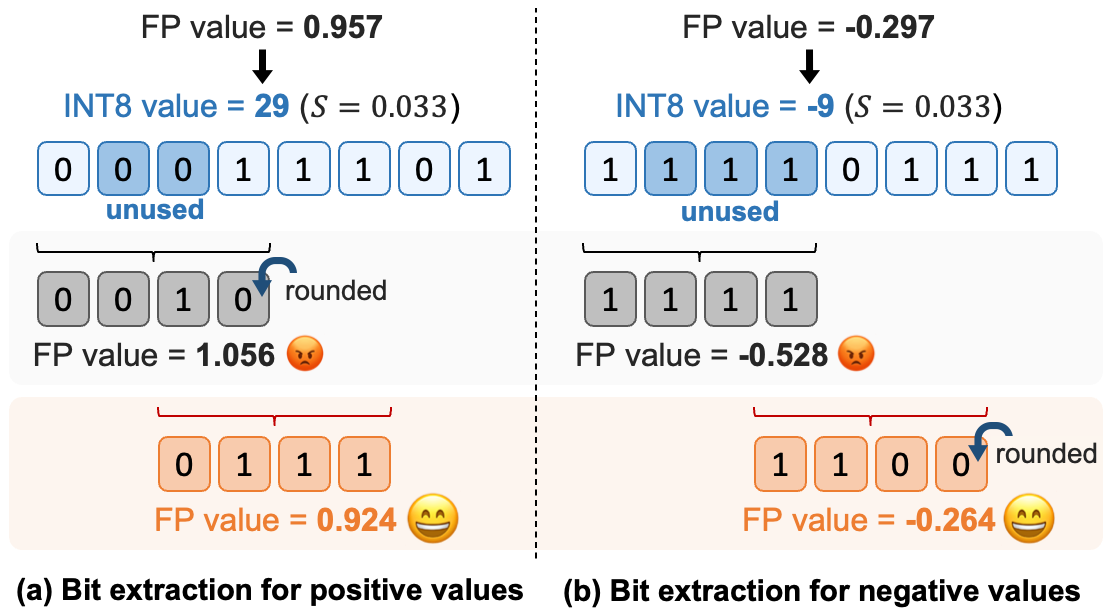}
\caption{Converting an 8-bit quantization value (top) into 4-bit in FlexiQ (bottom) and uniform quantization (middle).}
\label{fig:BitExtracting}
\end{figure}

Figure\,\ref{fig:BitExtracting} illustrates an example when we apply lowering from 8-bit to 4-bit quantization. The full-precision value of 0.957 is represented as 29 in 8-bit quantization. If this value comes from a feature channel where the maximum is less than 32, the highest two bits, except the sign bit, are not used. A naive conversion to 4-bit takes the highest four bits, resulting in a 10\% error, as shown in Figure\,\ref{fig:BitExtracting}(a) in the middle. In FlexiQ, we extract the third to sixth bits, excluding the unused bits (as the third bit is always equivalent to the sign bit), resulting in less than a 4\% error with six effective bits instead of four. The bit extraction logic is simple with shifting and storing low four bits and can be efficiently implemented on NPUs and GPUs as we explain in Section\,\ref{sec:npu-gpu-impl}.

The figure also illustrates an example on the right where the (absolute) maximum value is set by a negative value. For the 8-bit quantization value of -9 (mapping to full-precision -0.297), the recorded minimum value is larger than -16, and thus the three most significant bits are unused. Similarly to the previous example, we extract the fourth to seventh bits, as the fourth bit always equals to the sign bit.

When applying our bit-lowering method, we statically assign the extraction position for each feature channel in advance and extract bits from this pre-set position at runtime. Optionally, the extraction position can be adjusted dynamically by identifying unused bits at runtime. This is done by performing a bitwise OR operation across values within the same channel group to identify the highest unset bit, which determines where to extract the bits and how to shift them when accumulating with high-bitwidth computation results. This dynamic approach is particularly effective in improving accuracy when the ratio of low-bitwidth computation is high. The overhead is modest, measured at 2--5\% of the convolution or linear operation cost. We evaluate both the overhead and the resulting accuracy improvements in our experiments.

\begin{figure}[t]
\centering
\includegraphics[width=0.85\linewidth]{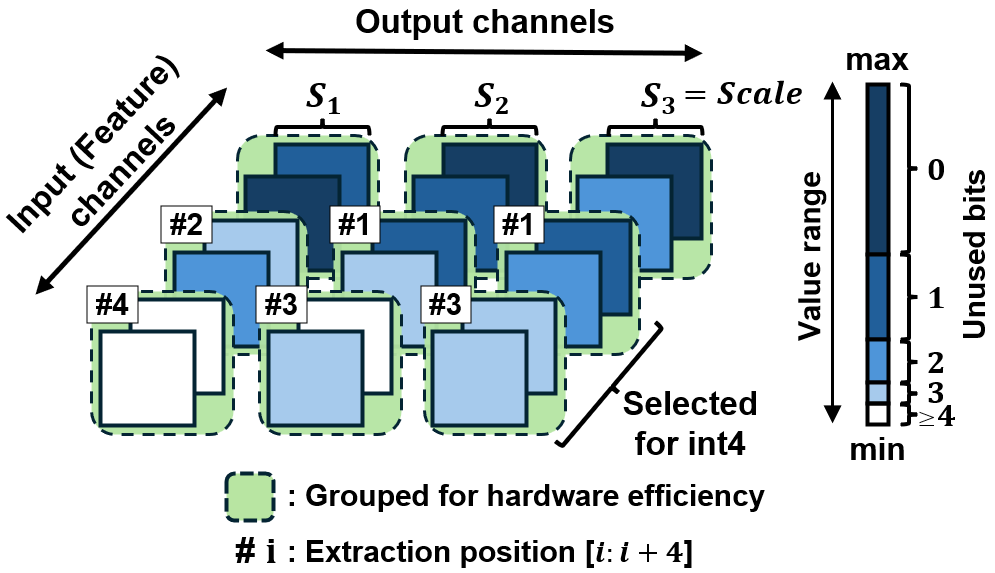}
\caption{Applying FlexiQ with different scale factors used for weight parameters in the output channel dimension. Channels selected for 4-bit computation in FlexiQ may use different bit extraction positions, denoted by \#1--\#4.}
\label{fig:ChannelQuant}
\end{figure}

Since we apply bit-lowering at the feature channel level, our technique is compatible with channel-wise quantization, which assigns different scale factors to weight parameters across the output channel dimension. Figure\,\ref{fig:ChannelQuant} shows this compatibility using a convolution operation as an example. Different scale factors ($S_1$–$S_3$) are assigned to output channels following the channel-wise quantization scheme. FlexiQ applies low-bitwidth computation to all parameters along the selected feature channel dimension (each with its corresponding scale factor $S_1$–$S_3$), using independently chosen bit extraction positions across output channels. This ensures compatibility with channel-wise quantization.

\subsection{Selecting Low-Bitwidth Channels}

When applying low-bitwidth quantization to a subset of feature channels, we prioritize selecting those that are likely to produce minimal quantization error. To achieve this, we examine the value ranges of the feature channels in each convolution or linear layer. For the channels with small ranges, our bit extraction ensures that low-bitwidth computation adds only small quantization errors.

We begin the selection by estimating the value ranges of the activation channels using sampled calibration data. For each activation channel, we examine the corresponding weight parameters' channels across the output channel dimension to determine the maximum value range. We then calculate an error estimation score by multiplying this maximum value range with the activation's range. Our bit extraction method is designed to ensure that higher scores correspond to higher quantization errors.

With the error estimation scores, we apply our evolutionary algorithm to select feature channels for a predetermined percentage of low-bitwidth computation. The algorithm, explained in Algorithm\,\ref{alg:evolutionary}, prioritizes the selection of the channels with low error scores, while also accounting for inter-layer interactions and the resulting error amplification. To optimize the selection efficiently, we use domain knowledge to design the chromosome encoding and mutation/crossover, as described below.

\begin{itemize}[leftmargin=0.25cm, noitemsep, topsep=0pt]
\item \textbf{Chromosome:} Each feature channel selected for low-bitwidth computation is represented by a bit flag, where 1 indicates low-bitwidth computation and 0 indicates high-bitwidth computation. A sequence of these bit flags, ordered by layer and channel, constitutes a chromosome.
\item \textbf{Crossover:} For two chromosomes, a single crossover point is randomly selected along a layer boundary. The bits to the right of this point are swapped between the two chromosomes, producing two offspring chromosomes.
\item \textbf{Mutation:} For a given chromosome, each bit flag set to 1 is examined and flipped with a small probability (by default 1\%). To maintain the same total number of set flags, an equal number of unset flags are randomly flipped in the same layer with the probability inversely proportional to the score of the channel. If the number of bit flags set is larger (or smaller) than the target percentage due to the crossover operation, this operation sets (or unsets) the bit flags to maintain the target percentage.
\end{itemize}

The mutation operation explores optimal low-bitwidth channels within each layer, while the crossover operation determines the suitable ratio of low-bitwidth calculations across the model's layers, accounting for the error amplification from inter-layer interactions. Together, these operations aim to identify the best combinations of low-bitwidth parameters to achieve the highest accuracy for the target percentage.

With these operations, Algorithm\,\ref{alg:evolutionary} starts with $N$ seed chromosomes (randomly initialized with higher probabilities for channels with lower error scores and one initialized with a greedy selection in a layer, each assigned an equal low-bitwidth ratio). We measure their fitness using the L2 distance from the soft labels of the high-bitwidth quantization model to capture the additional error introduced by low-bitwidth quantization. Using an elitist selection strategy, $k$ chromosomes with the highest fitness are carried over to the next generation (by default $k$=2). Then we apply the crossover and mutation operations to top $r$ chromosomes to generate offspring chromosomes. By repeating this process across generations, we obtain the optimal combination of low-bitwidth parameters for the target percentage.

When running the algorithm, we apply the operations at the granularity of multiple channels, ensuring that NPUs and GPUs can efficiently execute low or high-bitwidth calculations, fully utilizing the compute resources. That is, during chromosome initialization or mutation, we ensure that the low-bitwidth channels in each layer are multiples of a specified value, determined by the underlying hardware architecture (details in Section\,\ref{sec:npu-gpu-impl}). For efficient dynamic low-bitwidth ratio adjustment, we ensure that the channels selected at smaller low-bitwidth percentages are included in higher percentages; then we store these channels contiguously in memory with our layout optimization, explained in Section\,\ref{sec:postprocessing}, thereby avoiding runtime channel reordering.

\begin{algorithm}[t]
\centering
\small
\caption{Finding Low-Bitwidth Feature Channels}
\begin{algorithmic}[1]
\REQUIRE sample dataset $S$, generation size $G$,
\newline \phantom{12345} population size $N$, elite size $k$, parent size $r$
\STATE $C = \text{initialize } N \text{ chromosomes}$
\FOR{generation in 1\,..\,$G$}
    \STATE $\theta_c, \theta_{high} \gets \text{get\_params}(C)$
    \STATE $scores \gets \text{Inference}(S; \theta_c, \theta_{high})$
    \STATE $elites \gets \text{TopK}(scores, k)$
    \STATE $parents \gets \text{TopK}(scores, r)$
    \STATE $offsprings \gets \text{Crossover}(parents, N-k)$
    \STATE $offsprings \gets \text{Mutate}(offsprings, 0.01)$
    \STATE $C \gets elites + offsprings$
\ENDFOR
\STATE $\theta_c, \theta_{high} \gets \text{get\_params}(C)$
\STATE $scores \gets \text{Inference}(S; \theta_c, \theta_{high})$
\STATE $C_{best} \gets \text{Top-1}(scores)$
\end{algorithmic}
\label{alg:evolutionary}
\end{algorithm}

\section{Post-processing for Layout Optimization}
\label{sec:postprocessing}

While the evolutionary algorithm inclusively select low-bitwidth channels at increasing ratio, the selected channels may not be stored contiguous in memory. This complicates mapping low- and high-bitwidth calculations to hardware units. Therefore, after the channel selection, we reorder the feature channels in all layers to ensure that same-bitwidth channels are stored contiguously in memory.

The memory layout optimization involves three steps:
\begin{enumerate}[leftmargin=0.25cm, noitemsep, topsep=0pt]
\item \textbf{Reordering Initial Layer Channels:} The input channels of the first layer are reordered so that channels with the same bitwidth are stored contiguously (for all low-bitwidth ratios). The weight parameters of this layer are reordered accordingly.
\item \textbf{Reordering Subsequent Layers:} For each following layer, the input channels are reordered by adjusting the previous layer's weight parameters in the output channel dimension. This process is repeated for all layers sequentially.
\item \textbf{Handling Residual Connections:} An operator is inserted to reorder the output channels of previous layers at runtime, ensuring they match the order of the subsequent layer's output channels.
\end{enumerate}

The first two steps are performed statically, with no runtime overhead. The reordering operators inserted in step 3 execute at runtime but it incurs minimal overhead; on GPUs it is similar to a simple memory copy operation and is overlapped with the layer executions. On NPUs, an operation with an outgoing residual connection (without down-sample convolutions) stores its output additionally in the target memory location with reordering, which incurs small overhead (3\% of the total execution time).

\section{Optional Finetuning with Specialized Loss Function}
\label{sec:qat}

While the quantization scheme described in Section\,\ref{sec:ptq} delivers high accuracy and efficiency, its performance can be further improved through finetuning. The main difficulty is improving the accuracy of predominantly or fully low-bitwidth models without compromising that of high-bitwidth models. To address this, we introduce a finetuning algorithm that incorporates a specialized loss function.

\noindent
{\bf Loss Function for Bitwidth Control.} During finetuning, we perform forward propagation twice: once with low-bitwidth, followed by high-bitwidth computation, using both results for backpropagation. In the backpropagation step, we apply a specialized loss function to ensure the weight parameters remain compatible with both low and high-bitwidth configurations. This loss function applies both hard and soft labels as follows:
\begin{gather}
L^k = L_{CE}(p(x;\theta_k)|\hat{y}) + L_{CE}(p(x;\theta_k)|p(x;\theta_{fp32})) \label{eq:3} \\
L_{Total} = \lambda \times L^{low} + (1 - \lambda) \times L^{high} \phantom{0123456} \label{eq:4}
\end{gather}

The first term of equation (\ref{eq:3}) is the cross-entropy loss measured with the hard labels for the inference with $k$-bit calculation. The second term is the distillation loss using soft labels from the full-precision model inference. We apply this loss function in equation (\ref{eq:4}) with $k$ set to low and high-bitwidth. By considering both low and high-bitwidth computations simultaneously, the loss function ensures high-bitwidth inference accuracy while effectively training for low-bitwidth inference. With the parameters trained in this manner, we use our evolutionary algorithm to select a subset of feature channels for low-bitwidth computation.

\begin{figure}[t]
\centering
\includegraphics[width=0.95\linewidth]{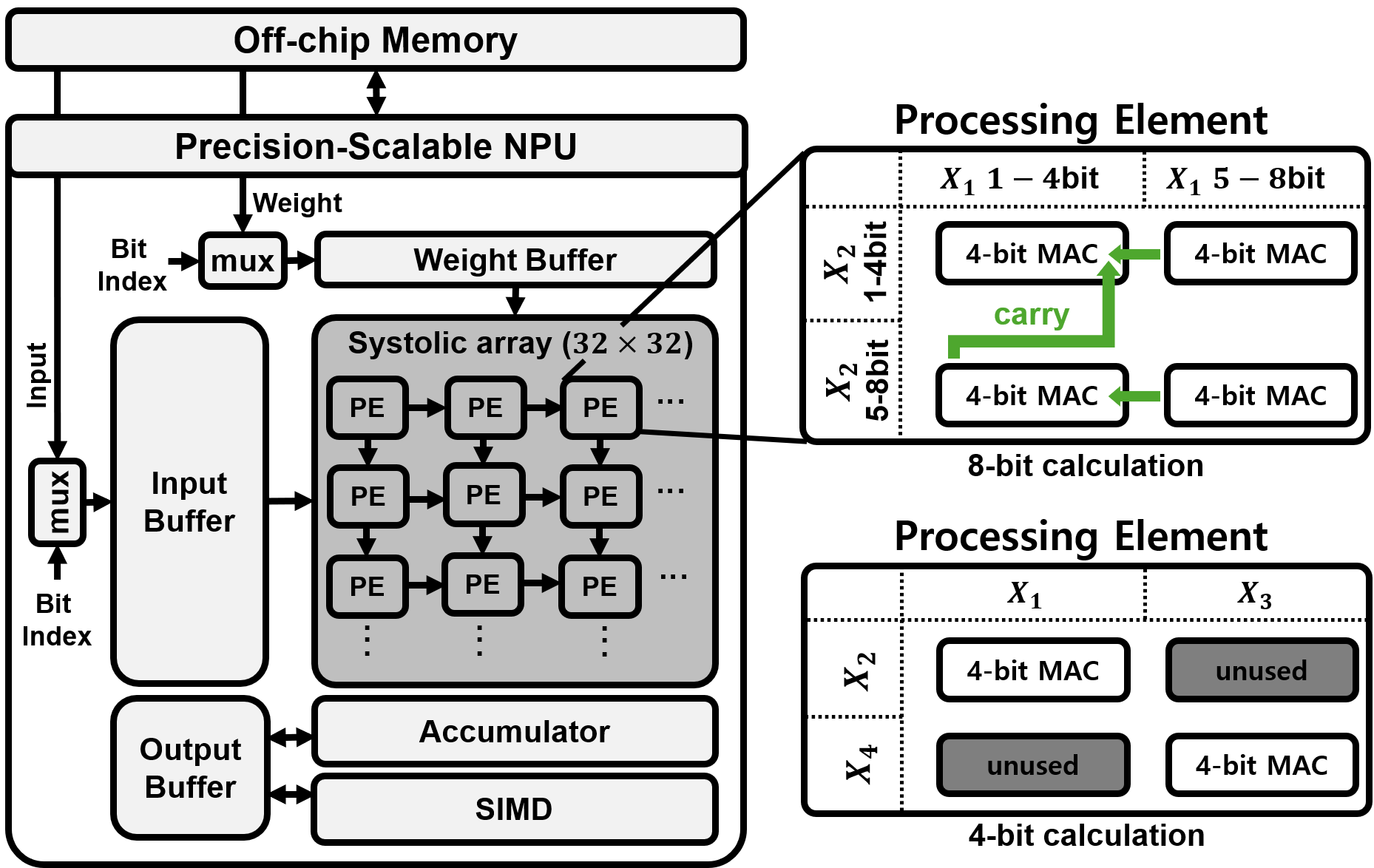}
\caption{Architecture of our 4/8-bit mixed-precision NPU.}
\label{fig:NPU}
\end{figure}

\section{NPU and GPU (CUDA) Implementation}
\label{sec:npu-gpu-impl}

We implemented FlexiQ's dynamic mixed-precision inference on both NPU and GPU. For NPU inference, we developed a custom NPU based on DNNWeaver v2\,\cite{sharma2016dnnweaver}, an open-source systolic array NPU. Since DNNWeaver does not inherently support mixed-precision calculations, we extended its design to include 4-/8-bit mixed-precision computation. 

Figure\,\ref{fig:NPU} shows the architecture of our NPU. Each Processing Element (PE) has four 4-bit Multiply-Accumulation (MAC) units, and the NPU has total of 1024 (32$\times$32) PEs. For 8-bit calculation, all four MAC units in each PE are used. For 4-bit calculations, two MAC units are used to perform two multiply-accumulation in parallel. The PEs are mapped such that the rows correspond to the input channels of the layer and the columns to the output channels. Thus, for 4-bit calculations, a group of sixty-four input channels is required to fully utilize all the PEs.

We apply weight-stationary dataflow, as it is widely adopted for NPUs. When computing a layer with mixed-bit precision, our NPU receives the number of 8-bit and 4-bit channels and switches the compute precision accordingly. In 4-bit compute mode, two 4-bit MAC results in each PE are accumulated to produce 64 16-bit values, which are bit-aligned based on the pre-set bit extraction positions and added to the 8-bit results in the accumulator. This precision switch incurs no pipeline bubbles, since 4-bit mode requires the same data bandwidth as 8-bit mode and the computation latency of the PEs remains unchanged.

\begin{figure}[t]
\centering
\includegraphics[width=0.77\linewidth]{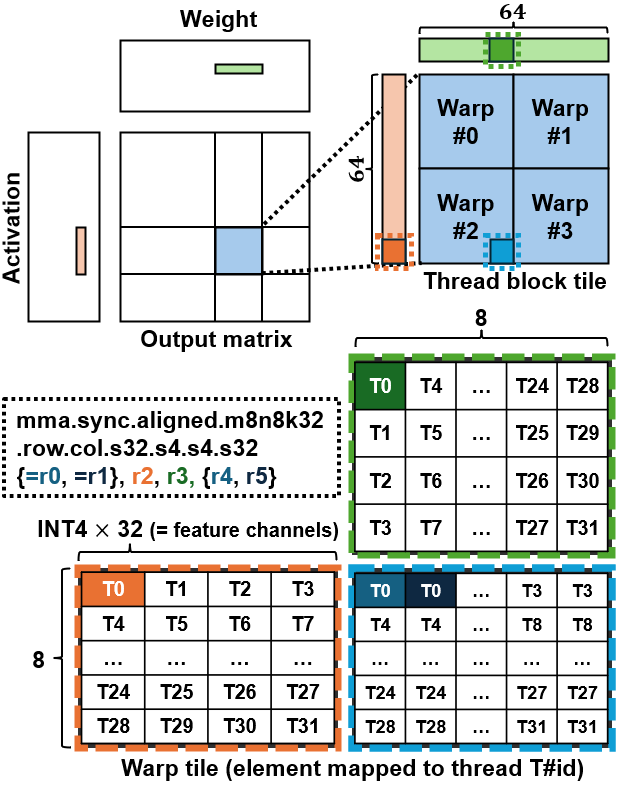}
\caption{Mapping GeMM computation to thread blocks and warps for MMA computation (black dotted box) in FlexiQ.}
\label{fig:GPU}
\end{figure}

For inference on GPUs, we implemented a dynamic mixed-bit GeMM kernel based on Nvidia's CUTLASS (v3.5.0), an open-source library for efficient GeMM operations\,\cite{thakkar_cutlass_2023}, and Atom\,\cite{zhao2024atom}, a mixed-precision GeMM kernel. As in their optimized implementations, we adopted a nested pipeline -- the outer pipeline handles data loading from global to shared memory and then to registers, while the inner pipeline overlaps register loading with multiply-accumulation. Our kernel uses Nvidia's MMA PTX instructions to exploit Tensor Cores' 4-bit compute capabilities (shown in Figure\,\ref{fig:GPU}).

For runtime efficiency, we execute multiply-accumulation of multiple feature channels with the MMA (Matrix Multiply-Accumulation) instruction. 
The instruction requires minimum 32 elements along the feature dimension for 4-bit multiply-accumulation\,\cite{googlesla}, and thus we set the feature group size for 4-bit computation to be 32 in GPUs. The MMA results are bit-shifted based on the group's bit extraction position and then combined with the 8-bit compute results.
While MMA computations run on Tensor Cores, the bit-shifting and mixed-precision accumulation are performed on CUDA Cores. These operations are pipelined, effectively hiding the overhead of bit-shifting (shown in Section\,\ref{sec:eval-npu/gpu}).

\noindent
\textbf{Low-bitwidth Ratio Adjustment.} 
Our mixed-precision GeMM kernel is based on post-processing layout optimization. This optimization guarantees that low-bit channels at lower ratios are subsets of those at higher ratios (e.g., all 25\% channels are included in 50\%). The kernel iterates over feature channels, performing bit extraction and 4-bit MMA instructions until reaching the last 4-bit channel, indicated by a global variable {\it max\_4bit\_ch}. It then switches to 8-bit computation for the remaining channels. Hence, adjusting the 4-bit ratio is achieved by updating {\it max\_4bit\_ch}.

\noindent
\textbf{Resource Consumption.} Since FlexiQ stores 8-bit model parameters to support dynamic 4-bit ratios, its memory footprint is equivalent to that of 8-bit models. The footprint could be reduced if only a narrower range of 4-bit ratios were supported (e.g., 50--100\% instead of 0--100\%). Because FlexiQ performs bit extraction at runtime, it incurs higher bandwidth usage than uniform 4-bit quantization. This overhead is mitigated by caching extracted 4-bit parameters, at the cost of additional memory. Furthermore, our kernel executes multiply-add operations in Tensor Cores, with bit-shifting and accumulation in CUDA Cores in a pipelined manner, making performance sensitive to the balance between CUDA and Tensor Core throughput (see Section\,\ref{sec:eval-npu/gpu}).

\begin{table}[t]
\centering
\small
\setlength\tabcolsep{1.5pt}
\caption{Evaluated models and training settings. \label{tab:models}}
\begin{tabular}{c|lr|c|c|c|c}
\hline\hline
\textbf{Dataset} & \multicolumn{1}{c}{\textbf{Model}} & (abbr.) & \begin{tabular}[c]{@{}c@{}}\footnotesize{\textbf{Calib.$^{\dagger}$}} \\ \footnotesize{\textbf{Data Size}}\end{tabular} & \begin{tabular}[c]{@{}c@{}}\footnotesize{\textbf{Finetune}}\\ \footnotesize{\textbf{Data Size}}\end{tabular} & \begin{tabular}[c]{@{}c@{}}\footnotesize{\textbf{Finetune}}\\ \footnotesize{\textbf{Epoch}}\end{tabular} & \footnotesize{\textbf{LR}}$^{\dagger}$ \\ \hline\hline

\footnotesize{Cifar10} & \multirow{2}{*}{ResNet-20} & \multirow{2}{*}{\scriptsize{(RNet20)}} & \multirow{2}{*}{128} & \multirow{6}{*}{100\%} & \multirow{2}{*}{50} & \multirow{2}{*}{\footnotesize{1e-3}} \\ \cline{1-1}
\footnotesize{Cifar100} &  &  &  &  &  &  \\ \cline{1-4} \cline{6-7} 
\multirow{10}{*}{\footnotesize{ImageNet}} & \footnotesize{ResNet-18} & \scriptsize{(RNet18)} & \multirow{4}{*}{256} &  & \multirow{4}{*}{20} & \multirow{4}{*}{\footnotesize{2e-4}} \\
 & \footnotesize{ResNet-34} & \scriptsize{(RNet34)} &  &  &  &  \\
 & \footnotesize{ResNet-50} & \scriptsize{(RNet50)} &  &  &  &  \\
 & \footnotesize{MobileNetV2} & \scriptsize{(MNetV2)} &  &  &  &  \\ \cline{2-7} 
 & \footnotesize{ViT Small} & \scriptsize{(ViT-S)} & \multirow{6}{*}{256} & \multirow{6}{*}{1\%} & \multirow{6}{*}{40} & \multirow{6}{*}{\footnotesize{1e-4}} \\
 & \footnotesize{ViT Base} & \scriptsize{(ViT-B)} &  &  &  &  \\
 & \footnotesize{DeiT Small} & \scriptsize{(DeiT-S)} &  &  &  &  \\
 & \footnotesize{DeiT Base} & \scriptsize{(DeiT-B)} &  &  &  &  \\
 & \footnotesize{SwinT Small} & \scriptsize{(Swin-S)} &  &  &  &  \\
 & \footnotesize{SwinT Base} & \scriptsize{(Swin-B)} &  &  &  &  \\ \hline\hline
\end{tabular}
\\\raggedleft\scriptsize$^{\dagger}$ Calibration Data Size \quad $^{\ddagger}$ Learning Rate \phantom{000}
\end{table}

\noindent
\textbf{Supporting Lower Precisions on NPU.} 
Our custom NPU supports 4/8-bit mixed-precision inference and can be extended to 2-bit computation by splitting each 4-bit MAC into two 2-bit MACs, enabling four parallel 2-bit MACs per 8-bit PE. This extension, however, creates a trade-off between hardware utilization and quantization granularity. Fully utilizing the 32$\times$32 systolic array in 2-bit mode requires a minimum channel group size of 128 (32 PEs $\times$ 4 MACs/PE), but grouping so many channels can reduce model accuracy. To mitigate this, spatial PE multiplexing\,\cite{lee2021dataflow} can partition the PE array to perform 2-bit and 8-bit computation concurrently. 

\section{Evaluation}
\label{sec:eval}

This section evaluates the trade-offs between latency and accuracy in deep learning models quantized with FlexiQ. We first describe the evaluation setup and then present the experimental results.

\subsection{Evaluation Setup}
\label{sec:eval-setup}

\begin{table*}[th]
\centering
\small
\setlength\tabcolsep{3pt}
\renewcommand{\arraystretch}{0.80} 
\caption{Accuracy (\%) of FlexiQ's 4/8-bit mixed-precision models. The models have 0--100\% of channel parameters quantized in 4-bit (specified at the header). The baselines are Uniform INT4 and INT8 channel-wise quantization. The accuracy of FlexiQ with 0\% 4-bit ratio equals that of INT8.\label{tab:mp-accuracy}} 
\begin{tabular}{l@{\hspace{1em}}c cccc c@{\hspace{1em}}c cccc c@{\hspace{1em}}c}
\toprule
& \textbf{baseline} & \multicolumn{5}{c}{\textbf{FlexiQ}} & \textbf{baseline} & \multicolumn{5}{c}{\textbf{FlexiQ (+ finetuned)}} & \\
\cmidrule(lr){2-2} \cmidrule(lr){3-7} \cmidrule(lr){8-8} \cmidrule(lr){9-13}
\textbf{Model} & \textbf{Uniform} & \textbf{100\%} & \textbf{75\%} & \textbf{50\%} & \textbf{25\%} & \textbf{Uniform} & \textbf{Uniform} & \textbf{100\%} & \textbf{75\%} & \textbf{50\%} & \textbf{25\%} & \textbf{Uniform} & \textbf{Full-Prec.} \\
& \textbf{INT4} & \small{(4-bit)}$^{\dagger}$ & \small{(5-bit)} & \small{(6-bit)} & \small{(7-bit)} & \textbf{INT8} \small{(=0\%)} & \textbf{INT4} & \small{(4-bit)} & \small{(5-bit)} & \small{(6-bit)} & \small{(7-bit)} & \textbf{INT8}$^{\ddagger}$ & \\
\midrule
ViT-S & \phantom{0}0.33 & 77.09 & 78.79 & 80.06 & 80.63 & 81.10 & 54.72 & 78.64 & 79.54 & 80.27 & 80.55 & 81.41 & 81.39 \\
ViT-B & \phantom{0}0.13 & 80.23 & 81.38 & 81.73 & 81.86 & 81.40 & 73.48 & 83.81 & 84.42 & 84.67 & 84.63 & 84.72 & 84.54 \\
Swin-S & \phantom{0}0.70 & 81.75 & 82.72 & 82.87 & 83.12 & 83.14 & 77.68 & 82.52 & 83.07 & 83.23 & 83.36 & 83.43 & 83.23 \\
Swin-B & \phantom{0}0.34 & 83.97 & 84.55 & 84.68 & 84.84 & 84.92 & 77.83 & 84.36 & 84.70 & 84.79 & 84.87 & 84.97 & 85.27 \\
DeiT-S & \phantom{0}0.97 & 76.49 & 77.96 & 78.39 & 78.62 & 78.74 & 68.23 & 78.42 & 79.06 & 79.35 & 79.59 & 79.65 & 79.85 \\
DeiT-B & \phantom{0}0.93 & 80.38 & 81.27 & 81.32 & 81.37 & 81.49 & 74.57 & 81.03 & 81.27 & 81.42 & 81.44 & 81.75 & 81.80 \\
\midrule
RNet18 & 51.03 & 66.80 & 68.81 & 69.25 & 69.47 & 69.62 & 68.16 & 68.35 & 69.43 & 69.52 & 69.95 & 70.01 & 69.76 \\
RNet34 & 64.88 & 71.21 & 72.40 & 72.96 & 73.23 & 73.31 & 72.06 & 72.62 & 73.08 & 73.38 & 73.50 & 73.44 & 73.30 \\
RNet50 & 65.99 & 73.27 & 75.37 & 75.72 & 75.98 & 76.13 & 74.37 & 75.67 & 76.24 & 76.44 & 76.57 & 76.53 & 76.14 \\
MNetV2 & 26.83 & 54.71 & 66.11 & 69.67 & 70.89 & 71.43 & 67.31 & 69.29 & 71.33 & 71.93 & 72.42 & 72.30 & 71.87 \\
\bottomrule
\end{tabular}
\\\raggedleft\footnotesize$^{\dagger}$ Average bitwidth (activation and weight parameters)  \quad $^{\ddagger}$ Finetuned with FlexiQ's loss function\phantom{000} 
\end{table*}

For the evaluation, we implemented the channel selection and training algorithm in PyTorch 2.4 using Python 3.10. The dynamic bit extraction and mixed-bit GeMM were implemented in CUDA 12.1 with latency evaluations conducted on Nvidia A6000.
We used five convolutional neural networks (CNNs) and six vision transformer models that are widely used in recent studies for evaluating quantization techniques\,\cite{he2016residual, HowardZCKWWAA17, touvron21a, Liu_2021_ICCV}. These models are also commonly employed as backbones for tasks such as object detection and diffusion\,\cite{fang2021you, zheng2021rethinking, peebles2023scalable, wang2023diffusion}. The evaluated CNNs are ResNet-18, ResNet-34, ResNet-50, and MobileNet-V2, while the transformer models are ViT-Small, ViT-Base, DeiT-Small, DeiT-Base, SwinT-Small, and SwinT-Base, as shown in Table\,\ref{tab:models}.

For these models, we used their publicly available pretrained versions from TorchVision\,\cite{torchvision2016} and the Hugging Face repository\,\cite{rwightman2021timm}, then quantized them with FlexiQ. 
When applying finetuning, we trained computer vision models for 20 epochs and transformer models for 40 epochs. We employed the widely used SGD optimizer for quantization training, starting with a learning rate of 1e-3 for the CIFAR dataset\,\cite{krizhevsky2009learning} and 1e-4 for the ImageNet dataset\,\cite{ImageNet}. A learning rate decay of 0.1 was applied every 10 epochs, with a weight decay of 1e-4. For the weight parameters, we used channel-wise quantization to minimize quantization errors. For activations, we determined the quantization ranges using the exponential moving average with the momentum of 0.99 across batches. For our specialized loss function we set $\lambda$ to be 0.5.

We used a combination of 4-bit and 8-bit computations for mixed-precision quantization. During feature channel selection for 4-bit quantization, we ensured that each layer’s selected channels were multiples of 32 on GPUs and 64 on the NPU. For our evolutionary selection algorithm, we set a population size of 50 and ran it for 50 generations. The mutation probability was 1\%, the elite size was 2, and the crossover parent size was 10.

\subsection{Overall Accuracy Across Low-Bitwidth Ratios}
\label{sec:eval-mp}

We evaluated the accuracy of models quantized by FlexiQ with 4/8-bit mixed-precision computations. As is conventional for low-bitwidth quantization, the first and last layers are computed in 8-bit precision. 
Similar to the other schemes we compared, we used 4-bit or 8-bit integer computation for linear and convolution operations, and 16-bit floating-point computation for other operations such as normalization and GELU.
For this evaluation, we varied the proportion of 4-bit parameters from 0\% to 100\% in increments of 25\%. For comparison, we applied uniform channel-wise quantization at 4-bit or 8-bit. We refer to these baselines as Uniform INT4 or Uniform INT8.

Table\,\ref{tab:mp-accuracy} summarizes the evaluation results. The accuracies of Uniform INT8 models closely match that of the full-precision model shown in the rightmost column. When half of the model’s computation is performed in 4-bit, the accuracy loss compared to Uniform INT8 remains under 0.4\% for most models with or without finetuning (average 0.6\% loss compared to full-precision), except for ViT-Small (1.04\% loss) and MobileNet (for finetuned version, 1.76\% loss). 
When finetuning is applied, FlexiQ’s 100\% 4-bit model achieves an average accuracy improvement of 6.6\% over the Uniform INT4 model (4.7\% excluding ViT-S). This shows that FlexiQ effectively manages the trade-off between accuracy and computational cost. Moreover, most of 100\% 4-bit models in FlexiQ incur less than 3\% loss due to its effective bit-lowering method.

As the proportion of 4-bit parameters increases from 0\% to 100\% in the evaluated models, accuracy declines gradually at first, then drops more sharply when approaching a fully 4-bit configuration. This suggests that performing even a small fraction of the calculations in 8-bit can noticeably boost accuracy compared to 100\% 4-bit models. Hence, FlexiQ achieves a balanced trade-off between maintaining high accuracy and lowering the average bitwidth, leading to more efficient and effective model performance.

\begin{figure}[t]
\centering
\includegraphics[width=0.95\linewidth]{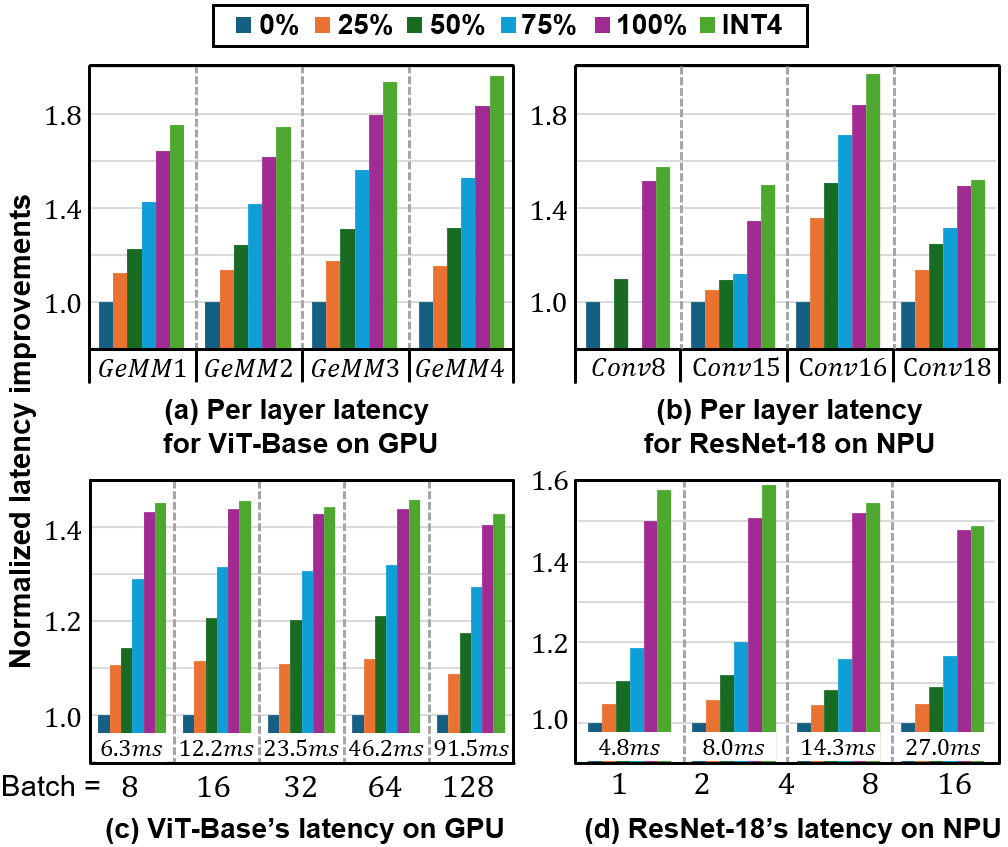}
\caption{Latency for ViT-Base on GPU (left, A6000) and ResNet-18 on NPU. The top plots are the latencies of GeMM and convolutions. The bottom plots are the model inference latencies (8-bit model latencies shown on x-axis).}
\label{fig:gpu-npu-latency}
\end{figure}

\begin{figure}[t]
\centering
\includegraphics[width=0.90\linewidth]{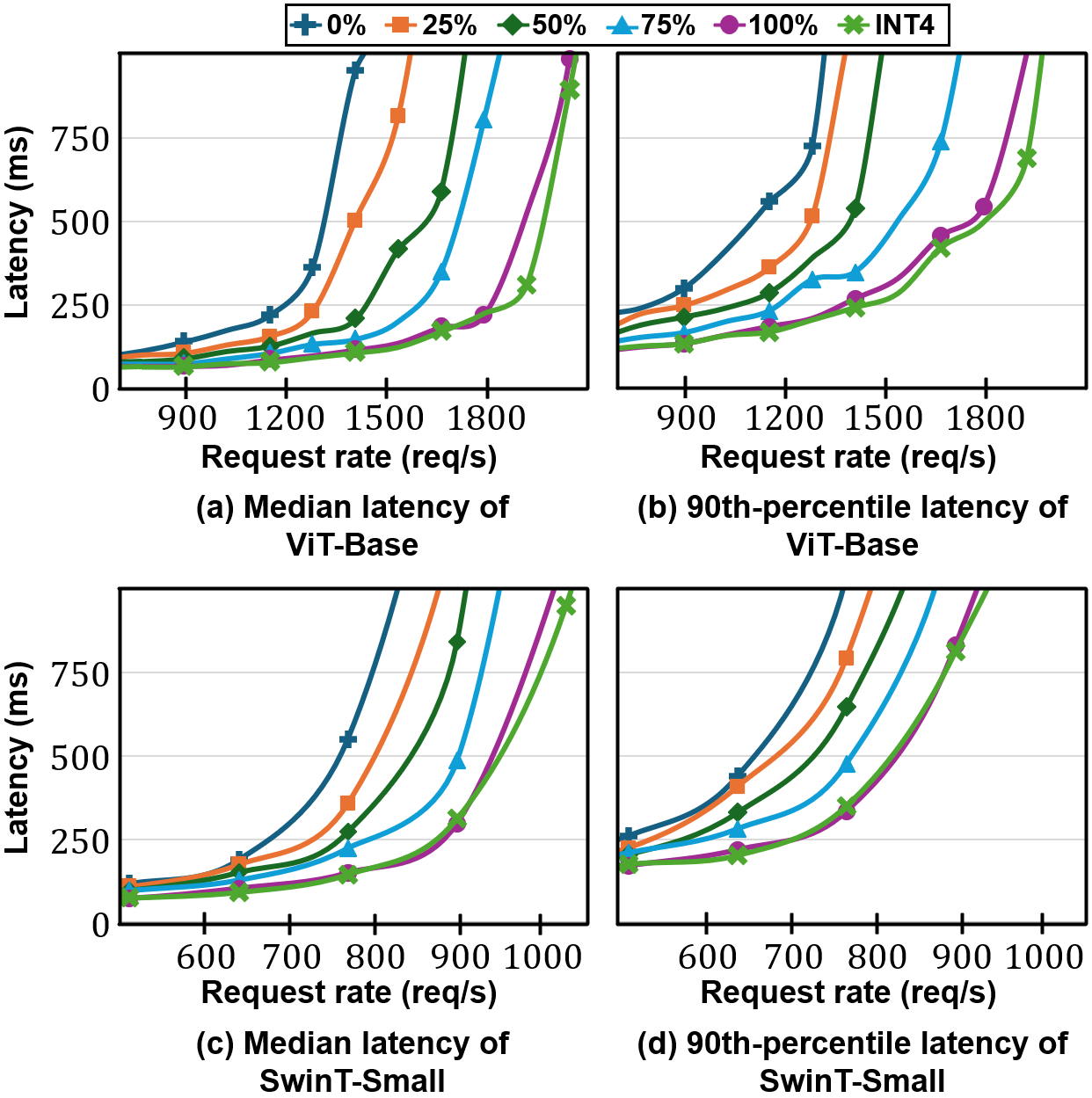}
\caption{Median and 90th-percentile latencies of FlexiQ (0–100\% 4-bit ratios) and INT4/INT8 quantization for ViT-Base and SwinT-Small on an A6000 GPU. Requests are generated using a Poisson distribution.}
\label{fig:e2e-model-latency}
\end{figure}

\subsection{Inference Latency on NPU and GPU}
\label{sec:eval-npu/gpu}

This section evaluates inference latency on our mixed-precision NPU and our CUDA kernels on GPUs when running models quantized with FlexiQ. We measure both model- and layer-level latencies, as well as end-to-end request response times -- including queuing delays -- in real-world serving scenarios.

We first report the model- and layer-level latencies on NPU and GPU. For the experiments, the ratio of 4-bit computation increased from 0\% to 100\%. Figure\,\ref{fig:gpu-npu-latency} shows the results for ViT-Base on GPU and ResNet-18 on NPU that are representative of other models not shown. For the ResNet-18 latencies on NPU, the first layer is excluded as its execution with only three channels is not suited for weight-stationary parallelism (assuming it runs on the CPU or other devices).
On both platforms, higher 4-bit computation ratios improved inference latency, almost proportionally on GPU.

Specifically for ViT on GPU, the our mixed-precision GeMM kernel (supporting dynamic 4-bit ratio adjustment) with 100\% 4-bit computation runs 6\% slower than the INT4 baseline. However, because other operations in the transformer, such as attention, normalization, and GELU, are computed in 16-bit floating point, our model latency is on par with INT4 inference, achieving a 1.43$\times$ speedup compared to 8-bit computation. For ResNet on NPU, the small batch size results in some operations being closer to memory-bound, yet they show a modest latency-precision trade-off. In detail, runtime channel reordering incurs a 3\% overhead, and loading 8-bit tensors instead of 4-bit ones adds an additional 1–2\%.

To assess the impact of FlexiQ’s latency improvements on end-to-end request latency, including queuing delays, we evaluated it under real-world inference serving scenarios. We generated inference requests following a Poisson distribution with average rates ranging from 100 to 3000 requests per second and measured the resulting response times. Figure\,\ref{fig:e2e-model-latency} reports the median and 90th-percentile latencies for FlexiQ with 25--100\% 4-bit ratios and the baseline using INT4 and INT8 computation for a subset of transformer models; other models, including DeiT-S and DeiT-B, are omitted due to space limitations but show similar trends. FlexiQ with 100\% 4-bit inference achieves the latencies close to those of INT4 computation and supports 1.57$\times$ higher request rates than INT8 while maintaining similar 90th-percentile latency.

\begin{figure}[t]
\centering
\includegraphics[width=1.0\linewidth]{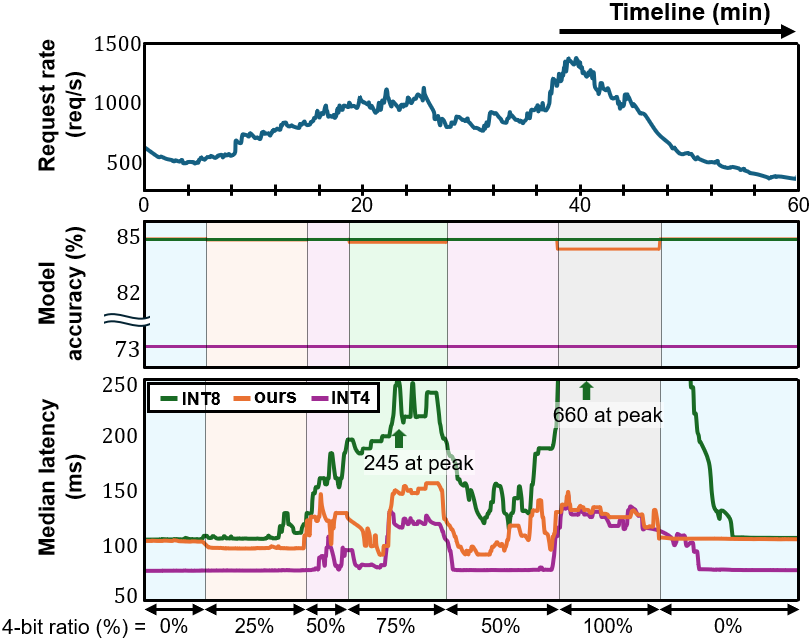}
\caption{Median latencies of FlexiQ and INT4/INT8 quantization for ViT-Base on an A6000 GPU under real-world fluctuating request traces.}
\label{fig:runtime-latency}
\end{figure}

\noindent
\textbf{Adapting to Workload Fluctuation.} We further evaluated FlexiQ’s dynamic low-bitwidth ratio adjustment using real-world inference traces. Following request rate fluctuation patterns observed in Azure traces\,\cite{azurepublicdataset, stojkovic2025dynamollm}, we generated requests with varying average rates, setting the peak rate to three times the minimum. At runtime, FlexiQ monitors the request rate and increases the 4-bit ratio by 25\% whenever the profiled latency (in Figure\,\ref{fig:e2e-model-latency}) corresponding to the current rate exceeds a predefined threshold.

Figure\,\ref{fig:runtime-latency} shows representative results. As the request rate fluctuates between 500 and 1500 requests per second, FlexiQ maintains stable median latencies between 100 and 150 ms, while INT8 inference latency rises as high as 660 ms. At peak request rates, FlexiQ’s latency closely matched that of INT4 inference. Notably, FlexiQ achieved an average accuracy of 84.64\%, nearly identical to INT8's 84.72\%.

\begin{table}[t]
\centering
\small
\setlength\tabcolsep{6pt}
\renewcommand{\arraystretch}{0.8} 
\caption{End-to-end latency (ms) of ViT-Base under different quantization and deployment frameworks.}
\label{tab:e2e-latency-frameworks}
\begin{tabular}{lcccc}
\toprule
& \multicolumn{4}{c}{\textbf{Batch Size}} \\
\cmidrule(lr){2-5}
\textbf{Method} & \textbf{16} & \textbf{32} & \textbf{64} & \textbf{128} \\
\midrule
CUTLASS INT8 & 14.42 & 28.95 & 58.17 & 126.93 \\
TensorRT INT8 & 14.13 & 27.87 & 55.05 & 111.47 \\
Uniform INT8$^{\dagger}$ & 12.24 & 23.53 & 46.23 & \phantom{0}91.55 \\
\midrule
FlexiQ 100\% & \phantom{0}8.67 & 16.78 & 32.71 & \phantom{0}65.00 \\
\midrule
Uniform INT4$^{\dagger}$ & \phantom{0}8.64 & 16.65 & 32.61 & \phantom{0}64.74 \\
CUTLASS INT4 & 14.22 & 28.66 & 57.68 & 125.95 \\
TensorRT INT4$^{\ddagger}$ & 21.00 & 36.74 & 70.10 & 136.52 \\
\bottomrule
\end{tabular}
\\\raggedleft\footnotesize$^{\dagger}$ Our custom kernel \quad $^{\ddagger}$ Weight-only quantization\phantom{0123456}
\end{table}

\noindent
\textbf{Baseline Performance.} Table\,\ref{tab:e2e-latency-frameworks} reports the performance of our baseline implementations (uniform 4-bit and 8-bit quantization) against CUTLASS and TensorRT. The uniform int8 baseline outperforms both CUTLASS and TensorRT int8 implementations. For 4-bit, the uniform int4 baseline shows a substantial advantage due to limitations in the other frameworks. CUTLASS requires an additional transformation from its column-major output format to PyTorch’s row-major format, while TensorRT lacks full INT4 support and is instead evaluated with weight-only quantization.

\begin{table}[t]
\centering
\small
\setlength\tabcolsep{1.5pt}
\renewcommand{\arraystretch}{0.85} 
\caption{Latency (ms) of ViT-Base on various GPUs.}
\label{tab:e2e-latency-vitb-gpu}
\begin{tabular}{lcccccccc}
\toprule
& \multicolumn{4}{c}{\textbf{Batch Size: 16}} & \multicolumn{4}{c}{\textbf{Batch Size: 128}} \\
\cmidrule(lr){2-5} \cmidrule(lr){6-9}
& \multicolumn{2}{c}{Commodity} & \multicolumn{2}{c}{Datacenter} & \multicolumn{2}{c}{Commodity} & \multicolumn{2}{c}{Datacenter} \\
\cmidrule(lr){2-3} \cmidrule(lr){4-5} \cmidrule(lr){6-7} \cmidrule(lr){8-9}
\textbf{Method} & \textbf{3090} & \textbf{A6000} & \textbf{A100} & \textbf{L40S} & \textbf{3090} & \textbf{A6000} & \textbf{A100} & \textbf{L40S} \\
\midrule
INT8 & 12.51 & 12.24 & 11.85 & 6.40  & 85.85 & 91.55 & 79.60 & 53.13 \\
FlexiQ 25\% & 11.84 & 11.08 & 11.66 & 5.97  & 84.42 & 83.92 & 78.30 & 53.47 \\
FlexiQ 50\% & 11.12 & 10.17 & 10.79 & 5.60  & 79.16 & 77.73 & 73.59 & 50.74 \\
FlexiQ 75\% & 10.25 & \phantom{0}9.40 & \phantom{0}9.77  & 5.11  & 73.15 & 71.09 & 67.71 & 47.05 \\
FlexiQ 100\% & \phantom{0}8.64  & \phantom{0}8.67  & \phantom{0}8.21  & 4.46  & 62.34 & 65.00 & 58.75 & 40.92 \\
INT4 & \phantom{0}8.41  & \phantom{0}8.64  & \phantom{0}7.67  & 4.20  & 59.58 & 64.74 & 51.97 & 40.75 \\
\bottomrule
\end{tabular}
\end{table}

\noindent
\textbf{Evaluation Across Different GPUs.} 
We evaluated FlexiQ on GPUs spanning multiple generations and both commodity and datacenter devices (Nvidia 3090, A6000, A100, and L40S), with results shown in Table\,\ref{tab:e2e-latency-vitb-gpu}. FlexiQ achieves speedups relatively proportional to the 4-bit ratio across all GPUs except A100. 
This exception results from our mixed-precision kernel running multiply-add operations on Tensor Cores but bit-shifting and accumulation on CUDA Cores, where A100’s lower CUDA Core throughput limits overall performance.

\subsection{Other Multi-Precision Quantization}
\label{sec:eval-cmp}

We compare FlexiQ to other multi-precision quantization schemes that support runtime bitwidth adjustment. Specifically, we evaluated RobustQuant\,\cite{chmiel2020robust}, AnyPrecision\,\cite{yu2021any}, and PTMQ\,\cite{xu2024ptmq}, as they support runtime bitwidth adjustment with low overhead. 
For reference, we include the accuracy of HAWQv3\,\cite{yao2021hawq}, although it does not support runtime bitwidth adjustment. RobustQuant and AnyPrecision controls the bitwidth of all parameters while PTMQ takes a layer-wise approach and controls bitwidth at the layer level. To the best of our knowledge, PTMQ is the only multi-precision scheme that has reported results for vision transformers.

Table\,\ref{tab:multi-cmp} compares the accuracy of quantized models using average bitwidths of 4-bit, 6-bit, and 8-bit computation. For the compared schemes, we use the results reported by their respective authors. A dash (--) indicates that the accuracy for the given setting was not reported. To assess the impact of adding or dropping bits on performance, we report the relative accuracy compared to that of the full-precision model in the table (denoted by FP).

For both 4-bit and 6-bit average bitwidth settings, FlexiQ consistently achieved the highest relative accuracy by a significant margin. In convolutional models, it outperformed other schemes by up to 3.0\%, while in vision transformers, the accuracy improvement reached 4.0\%. By managing bitwidth at the feature level -- a finer granularity than other approaches -- FlexiQ achieves higher accuracy in mixed-precision computation.

\begin{table}[t]
\centering
\small
\setlength\tabcolsep{3pt}
\renewcommand{\arraystretch}{0.85} 
\caption{Comparing multi-precision adaptive quantization schemes with four and six average bitwidths. Relative accuracy (vs. full-precision model, denoted by FP) is shown.}
\begin{tabular}{lccrrr}
\toprule
\textbf{Method} & \textbf{Finetuned} & \textbf{4-bit}$^{\dagger}$ & \textbf{6-bit}$^{\dagger}$ & \textbf{8-bit} & \textbf{FP (\%)} \\
\midrule
\multicolumn{6}{l}{\textit{\textbf{ResNet-18}}} \\
PTMQ & X & -3.43 & -0.77 & -0.21 & 71.00 \\
FlexiQ (ours) & X & -2.96 & -0.51 & -0.14 & 69.76 \\
HAWQv3 & O & -3.02 & -0.97 & +0.09 & 71.47 \\
RobustQuant & O & -3.40 & -0.30 & \multicolumn{1}{c}{---} & 70.30 \\
FlexiQ (ours) & O & \textbf{-1.41} & \textbf{-0.24} & \textbf{+0.25} & 69.76 \\
\midrule
\multicolumn{6}{l}{\textit{\textbf{ResNet-50}}} \\
PTMQ & X & -2.69 & -0.51 & -0.10 & 76.62 \\
FlexiQ (ours) & X & -2.87 & -0.42 & -0.01 & 76.14 \\
HAWQv3 & O & -3.48 & -1.77 & -0.14 & 77.72 \\
RobustQuant & O & -2.00 & -0.10 & \multicolumn{1}{c}{---} & 76.30 \\
AnyPrecision & O & -0.79 & \multicolumn{1}{c}{---} & -0.56 & 74.63 \\
FlexiQ (ours) & O & \textbf{-0.47} & \textbf{+0.30} & \textbf{+0.39} & 76.14 \\
\midrule
\multicolumn{6}{l}{\textit{\textbf{ViT-B}}} \\
PTMQ & X & \multicolumn{1}{c}{---} & -6.84 & -5.42 & \multirow{3}{*}{84.54} \\
FlexiQ (ours) & X & -4.30 & -2.81 & -3.14 & \\
FlexiQ (ours) & O & \textbf{-0.73} & \textbf{+0.13} & \textbf{+0.18} & \\
\midrule
\multicolumn{6}{l}{\textit{\textbf{DeiT-S}}} \\
PTMQ & X & \multicolumn{1}{c}{---} & -1.11 & -0.32 & \multirow{3}{*}{79.85} \\
FlexiQ (ours) & X & -3.36 & -1.46 & -1.11 & \\
FlexiQ (ours) & O & \textbf{-1.43} & \textbf{-0.50} & \textbf{-0.20} & \\
\midrule
\multicolumn{6}{l}{\textit{\textbf{DeiT-B}}} \\
PTMQ & X & \multicolumn{1}{c}{---} & -0.99 & -0.26 & \multirow{3}{*}{81.80} \\
FlexiQ (ours) & X & -1.42 & -0.48 & -0.31 & \\
FlexiQ (ours) & O & \textbf{-0.77} & \textbf{-0.38} & \textbf{-0.05} & \\
\bottomrule
\end{tabular}
\\\raggedleft\footnotesize$^{\dagger}$ average bits of activation and weight \phantom{000}
\label{tab:multi-cmp}
\end{table}

\subsection{Analysis of Our Evolutionary Algorithm}
\label{sec:eval-evol}

We evaluate our evolutionary algorithm for generating model versions with increasing mixed-bitwidth ratios. Using the error scores of channels estimated from their value ranges, it identifies multiple model versions with increasing percentages of low-bitwidth channels. Their accuracy has been reported in the previous section. Here, we measure the execution time of the evolutionary algorithm. We also compare the accuracy of the resulting mixed-precision model versions with those produced by other selection algorithms.

\begin{figure}[t]
\centering
\includegraphics[width=1.0\linewidth]{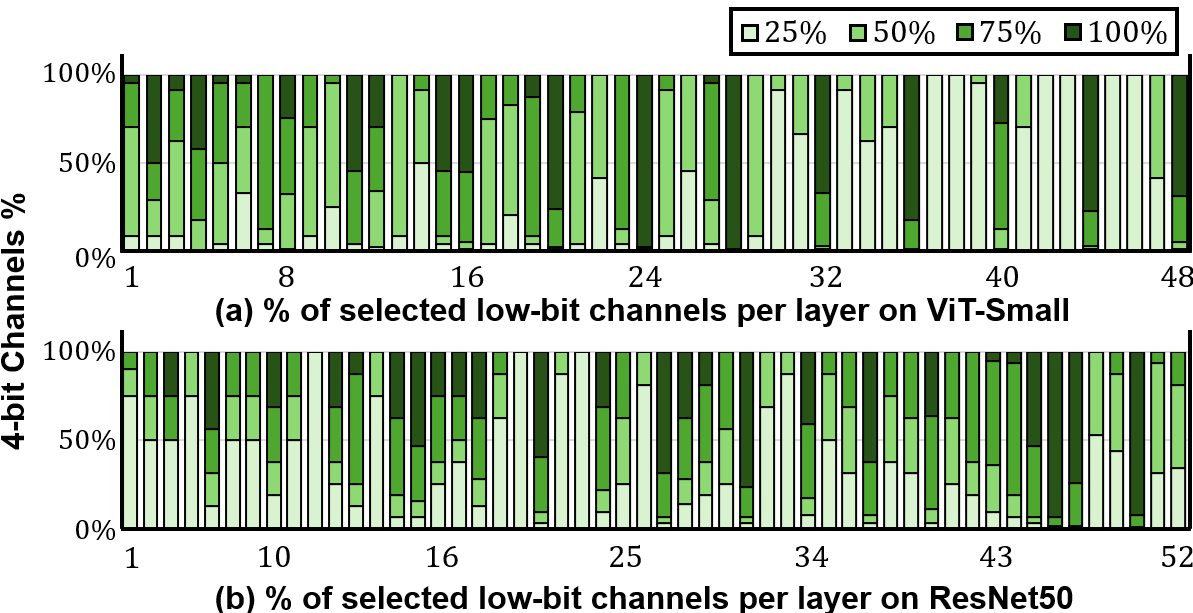}
\caption{Percentage of 4-bit channels in each layer of ViT-Small and ResNet-50 as the 4-bit ratio increases from 25\% to 100\%, as determined by our evolutionary algorithm.}
\label{fig:layer-4-bit-ratio}
\end{figure}

Our evolutionary selection begins by estimating the error scores and initializing seed chromosomes, a process that takes 2–10 seconds (MobileNet is the fastest, and SwinT-Base is the slowest). Next, running the evolutionary algorithm for 50 generations while measuring fitness on 256 input data for a given mixed-bitwidth ratio takes less than an hour, which remains within the typical PTQ processing time\,\cite{li2021brecq, xu2024ptmq}. Figure\,\ref{fig:layer-4-bit-ratio} presents an example selection results, showing the percentage of channels chosen for 4-bit computation.

\begin{figure}[t]
\centering
\includegraphics[width=1.0\linewidth]{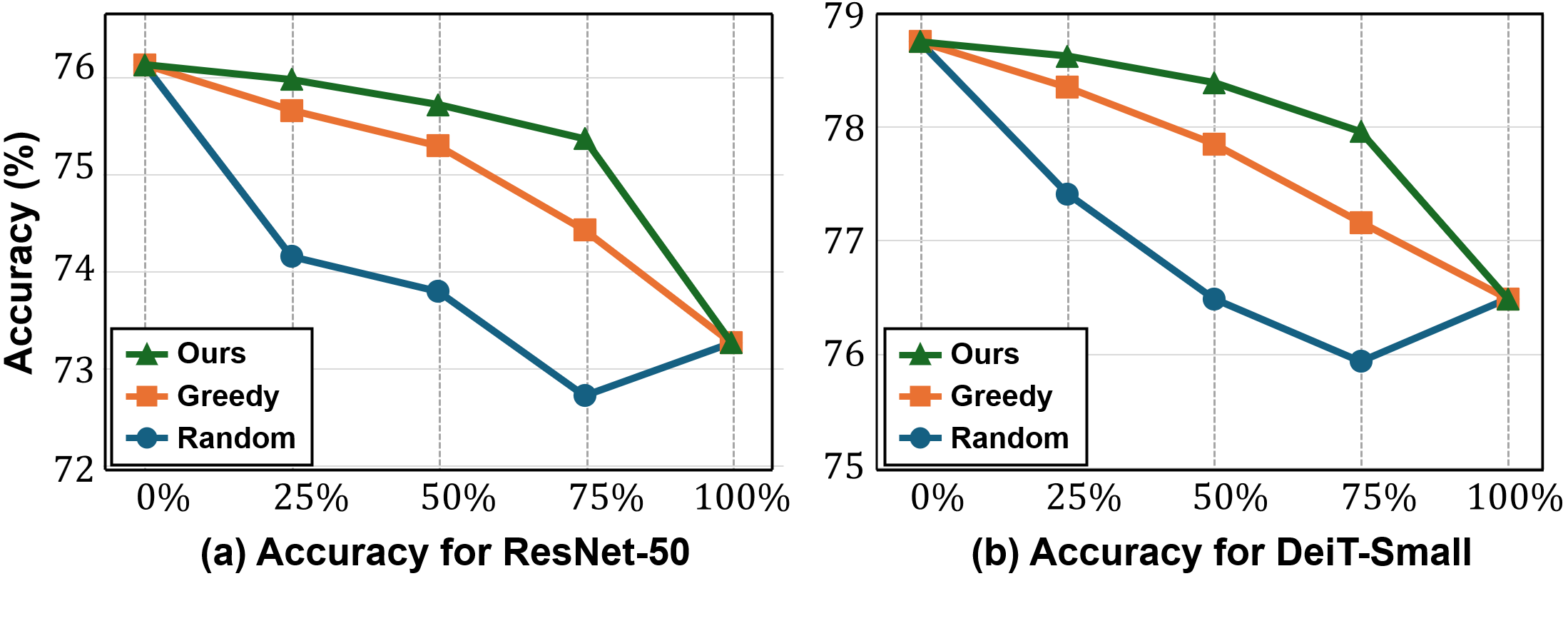}
\caption{Comparison of channel selection algorithms with 0--100\% 4-bit computation and 8-bit for the rest.}
\label{fig:adaptive}
\end{figure}

To evaluate the quality of the mixed-bit model versions that our algorithm generated, we implemented two selection algorithm, namely, random and greedy selection. As their names indicate, the former randomly selects feature channels and the latter greedily selects based on our error scores for 4-bit computation. Figure\,\ref{fig:adaptive} compares their results for both convolution- and transformer-based vision models. 

We observe that both the greedy selection and our evolutionary selection achieves 1.5–2\% higher accuracy than the random selection across the 25–75\% 4-bit quantization models. Compared to greedy selection, our approach gives a 0.2–1.0\% accuracy gain. Note that our (static) bit-lowering method is applied to all results; without it, random selection suffers a large accuracy loss (shown in Section\,\ref{sec:eval-ablation}). The results indicate that our evolutionary selection effectively finds low-bitwidth channels for mixed-bit computation.

\noindent
{\bf Overhead of adjusting precision at runtime.} 
The overhead for switching model versions is small, as parameters in each layer are aligned to ensure same-bitwidth parameters are stored contiguously for all low-bitwidth ratios. 
Switching between different 4-bit ratios therefore only requires updating a single variable for each layer that marks the maximum 4-bit channel index ({\it max\_4bit\_ch}, explained in Section\,\ref{sec:npu-gpu-impl}), which incurs negligible cost.  
On GPUs adjusting the ratio takes less than a few microseconds, and on our NPU, it requires under 0.3 microseconds to load instructions for the updated model into the instruction memory.

\noindent
\textbf{Manual Channel Selection.} We evaluated the effect of manually fixing a subset of channels to 8-bit computation on the evolutionary algorithm and resulting accuracies. Specifically, we randomly designated a portion of channels as 8-bit and then applied the algorithm to select 4-bit channels among the remainder. The results (omitted for space) show only minor accuracy degradation, roughly proportional to the fraction of fixed channels. For instance, with a target 50\% 4-bit ratio, fixing 40\% of channels to 8-bit reduced accuracy by just 0.20 percentage points.

\begin{figure}[t]
\centering
\includegraphics[width=0.95\linewidth]{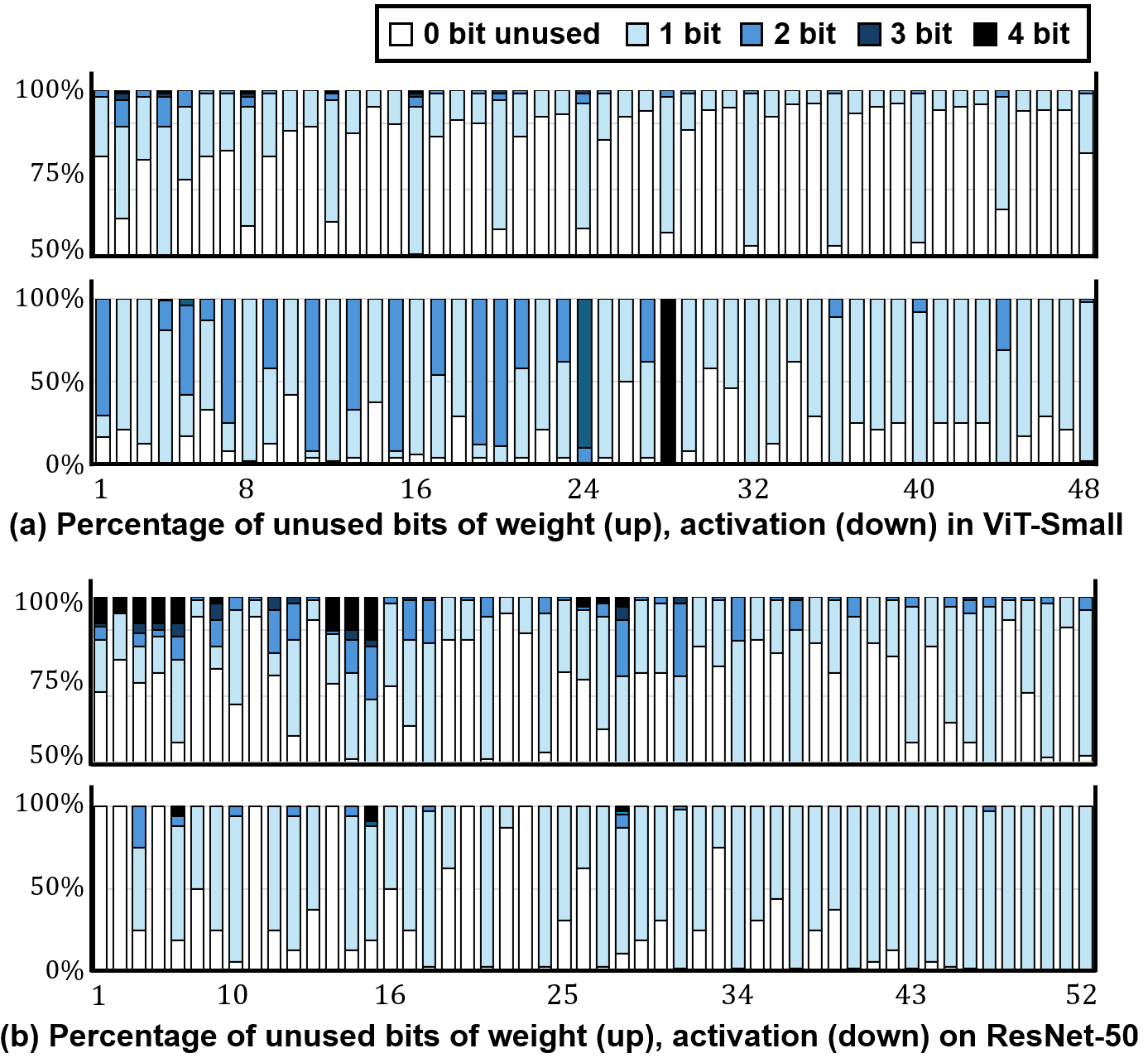}
\caption{The percentages of feature channels with 0--4 unused bits in each convolution/linear layer for ViT-Small (above) and ResNet-50 (below) measured with 1024 samples.}
\label{fig:unused-bits}
\end{figure}

\subsection{Analysis of Used and Unused Bits}
\label{sec:eval-bits}

This section examines the effectiveness of our range-based bit extraction method. First, we analyze the number of used and unused bits when applying our bit extraction with static extraction positions. Next, we measure the percentage of channels that become saturated under our bit extract method. Finally, we evaluate the overhead of dynamic extraction position and the resulting accuracy improvements.

For the analysis of used/unused bits, we presume the value ranges of the channels to cover 99\% of neuron values for each channel with 1024 sampled data. For each feature channel, we counted the number of unused bits with their value ranges. Figure\,\ref{fig:unused-bits} shows the results for ViT-Small and ResNet-50 that are representative of other models. In both models, 10--40\% of the weight parameter and activation channels have one or more unused bits with a large variation across the layers.

\begin{figure}
\centering
\includegraphics[width=1.0\linewidth]{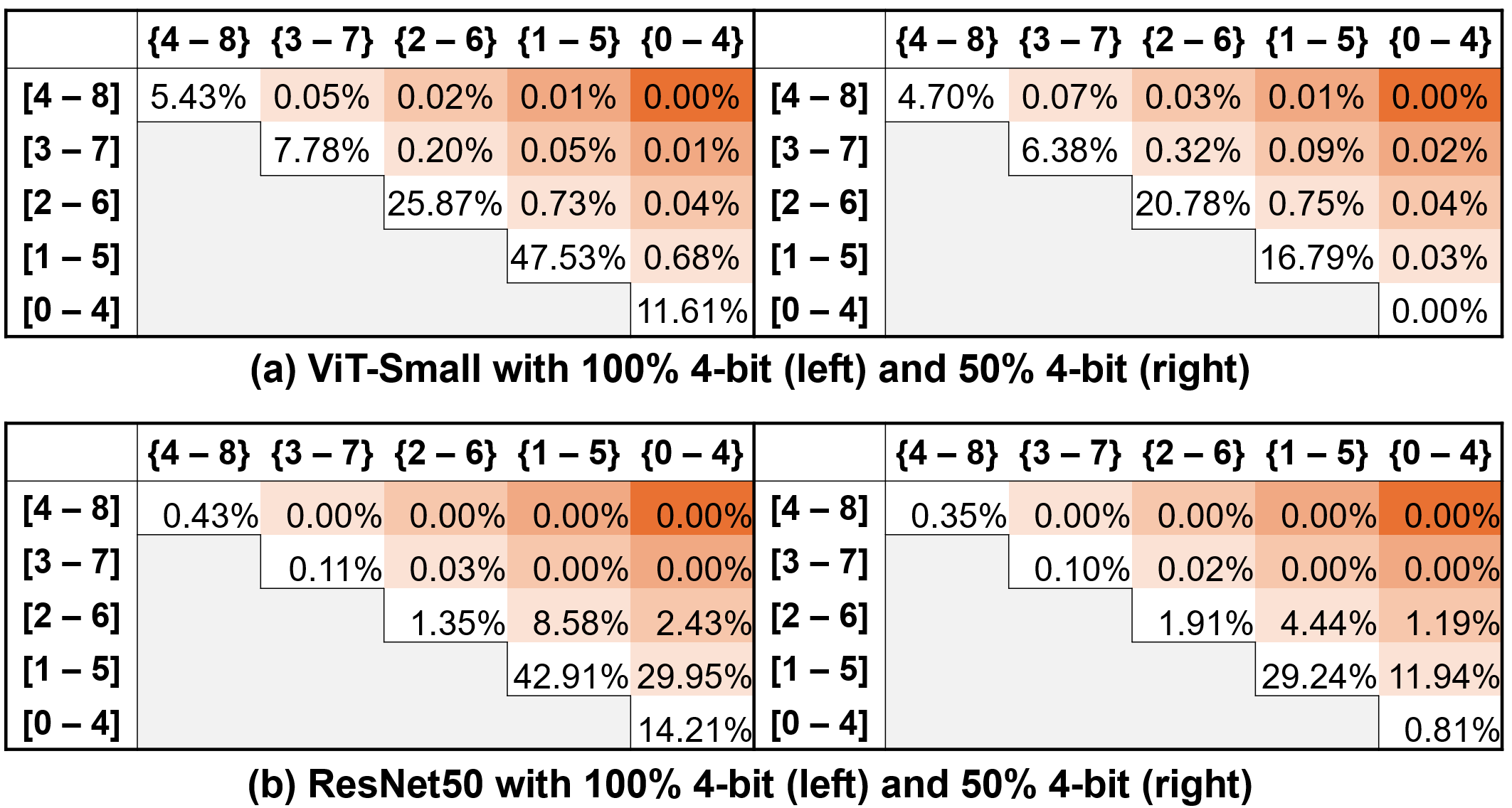}
\caption{Percentages of non-saturated (white) and saturated (orange) channels based on 1024 sampled data when applying statically identified bit-lowering positions. The notation [a–b] indicates the static bit extraction position from a to b, while \{c–d\} indicates the optimal extraction position from c to d.}
\label{fig:saturated-bits}
\end{figure}

Now we analyze saturated channels when applying our bit-lowering method using static bit extraction positions determined from calibration datasets.  Figure\,\ref{fig:saturated-bits} shows the distribution of saturated channels for ViT-Small and ResNet-50, representative of other models not shown due to space limit. Static positions, denoted by [a--b], indicate bit extraction from a to b-1, while \{c--d\} represents the optimal extraction positions that cover the used bits for the actual input data.

We noticed from the figure that vision transformers have only a small percentage of saturated channels when using static bit extraction positions. In contrast, convolutional models have a non-trivial number of saturated channels, but typically by one bit. As shown in the figure on the right, these saturated channels are given lower priority when FlexiQ selects 4-bit channels for 50\% of mixed 4-bit/8-bit computations.

\noindent
{\bf Finding extraction positions at runtime.} To reduce quantization errors in those saturated channels, we optionally support finding the highest unset bit for each channel at runtime by running a bitwise OR operation on the values in the same channel. The runtime overhead is measured to be 2--5\% of the corresponding convolution or linear operation. This gives an accuracy gain of 0.7--1.1\% (percent point) for the convolution models and 0.1--2.1\% for the vision transformers.

\subsection{Intra-Layer Mixed-Bit Computation}
\label{sec:eval-intra-layer-mp}

We compared the accuracy of computing a single layer in mixed 4-bit/8-bit precision against computing it entirely in 4-bit. To do this, we quantized each layer of a model with Uniform INT4 and INT8 while keeping the preceding layers in full precision. We also applied FlexiQ to the same layer with mixed 4-bit/8-bit quantization, varying the 4-bit proportion from 25\% to 100\% in increments of 25\%. We measured the L2 distance between the layer’s 8-bit output and both (1) the 4-bit output and (2) the output produced by FlexiQ.

\begin{figure}[t]
\centering
\includegraphics[width=0.95\linewidth]{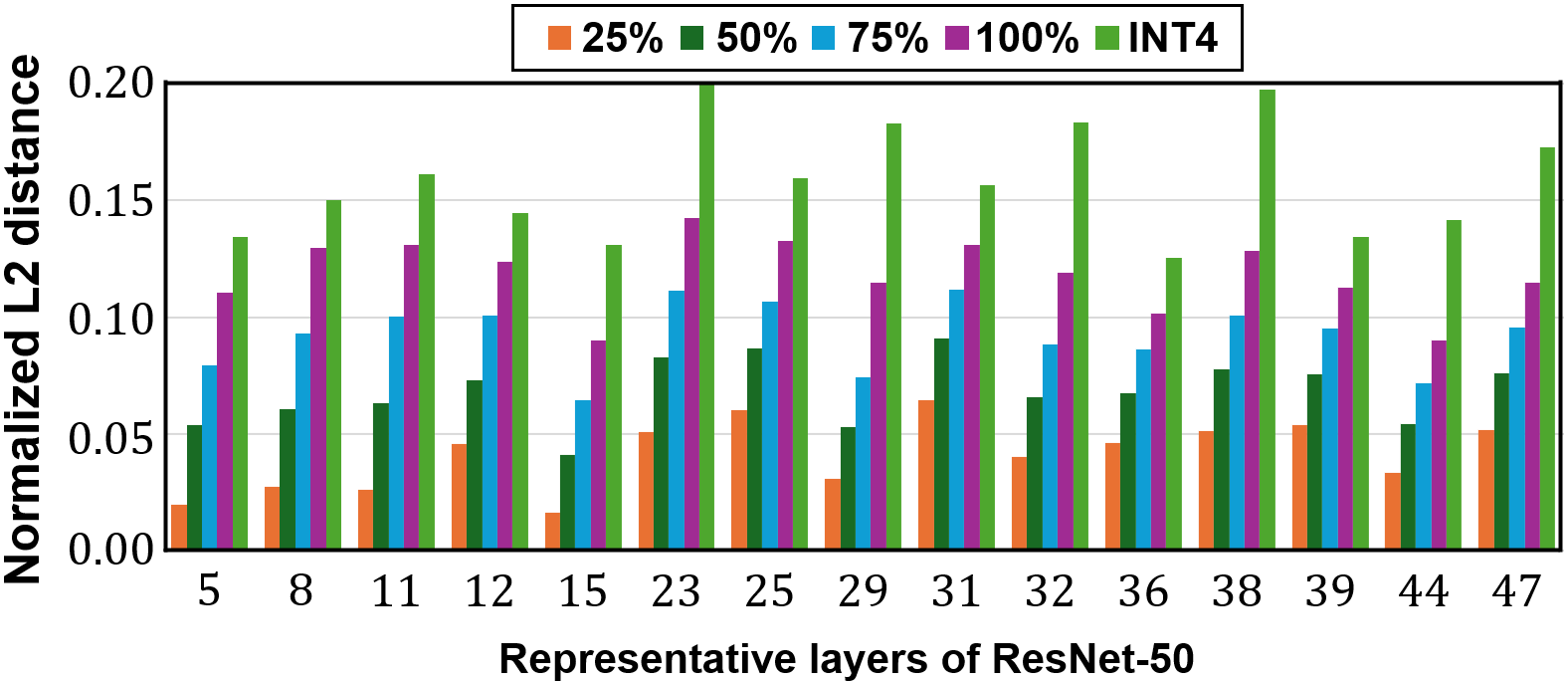}
\caption{L2 distance of output tensors in 4-bit quantization and FlexiQ (25--100\% 4/8 mixed-bit) to those in 8-bit quantization for each layer in ResNet-50. The distance is normalized to the L2 norm of the 8-bit output.
\label{fig:L2dist}}
\end{figure}

Figure\,\ref{fig:L2dist} shows the normalized L2 distance for 15 selected layers in ResNet-20 that are representative of the others. The distance for the Uniform INT4 quantization is large, with the minimum being 0.125, indicating a 12.5\% difference from the 8-bit results. Thus, even when a small number of layers are selected for 4-bit in layer-wise mixed-precision quantization, finetuning needs to adjust those layers' outputs by more than 12.5\% as well as the outputs of the subsequent layers.

For FlexiQ, all of the 25\% mixed-bit calculations (25\% in four-bit) result in less than an 4.7\% difference from the eight-bit calculation. For the 50\% mixed-bit calculations, half of the layers in the figure show less than an 7.4\% difference. This explains in part why FlexiQ achieves higher accuracy than other mixed-precision schemes. Note that the 100\% four-bit calculation in FlexiQ achieves higher precision by identifying unused bits and thus increasing the number of effective bits.

\begin{table}[t]
\centering
\small
\setlength\tabcolsep{1pt}
\renewcommand{\arraystretch}{0.80} 
\caption{Average errors relative to 8-bit inference in selected Q/K/V projection layers' output of ViT-Base under evolutionary (Evol.), greedy (Gred.), and random (Rand.) selection.}
\label{tab:layerwise_mae}
\begin{tabular}{c|ccc|ccc|ccc}
\toprule
\multicolumn{1}{c}{} & \multicolumn{3}{c}{\textbf{FlexiQ 25\%}} & \multicolumn{3}{c}{\textbf{FlexiQ 50\%}} & \multicolumn{3}{c}{\textbf{FlexiQ 75\%}} \\
\cmidrule(lr){2-4} \cmidrule(lr){5-7} \cmidrule(lr){8-10}
\footnotesize{\textbf{layer\#}} & \textit{Evol.} & \textit{Gred.} & \textit{Rand.} & \textit{Evol.} & \textit{Gred.} & \textit{Rand.} & \textit{Evol.} & \textit{Gred.} & \textit{Rand.} \\
\midrule
\#5 & 0.030 & 0.036 & 0.064 & 0.058 & 0.074 & 0.150 & 0.099 & 0.193 & 0.194 \\
\#9 & 0.070 & 0.079 & 0.120 & 0.108 & 0.127 & 0.207 & 0.168 & 0.256 & 0.248 \\
\midrule
\#21 & 0.130 & 0.141 & 0.204 & 0.204 & 0.226 & 0.312 & 0.286 & 0.366 & 0.361 \\
\#25 & 0.139 & 0.150 & 0.215 & 0.218 & 0.239 & 0.324 & 0.299 & 0.376 & 0.371 \\
\midrule
\#37 & 0.202 & 0.217 & 0.312 & 0.306 & 0.332 & 0.443 & 0.405 & 0.497 & 0.503 \\
\#41 & 0.246 & 0.262 & 0.375 & 0.361 & 0.397 & 0.522 & 0.478 & 0.584 & 0.591 \\
\bottomrule
\end{tabular}
\end{table}

\subsection{Layer-wise Error Analysis of Channel Selection}

The evolutionary selection algorithm accounts for inter-layer interactions and the resulting error amplification. We evaluated its layer-wise errors against greedy and random channel selection by measuring the L1 distance between 4/8-bit mixed-precision inference and 8-bit-only inference for each layer. Table\,\ref{tab:layerwise_mae} reports results on ViT-Base for six Q/K/V projection layers from the beginning, middle, and end of the model (other layers and models show similar trends and are omitted for brevity). The gap between evolutionary and greedy selection widens in deeper layers, indicating that evolutionary selection explicitly considers error amplification, whereas greedy selection considers only unused bits. The gap is also larger at higher 4-bit ratios (e.g., 75\% vs. 25\%). Overall, the evolutionary algorithm effectively mitigates error amplification by modeling inter-layer dependencies, yielding lower errors.

\subsection{Ablation Study}
\label{sec:eval-ablation}

We conducted an ablation study to examine the accuracy impact of our proposed techniques: 1) bit-lowering based on value ranges, 2) low-bitwidth channel prioritization, 3) evolutionary channel selection, 4) dynamic adjustment of bit extraction positions, and 5) finetuning.

For this study, we evaluated FlexiQ with applying the above five optimizations cumulatively. Specifically, the baseline, {\it Random}, randomly selects 4-bit channels and always extracts the highest four bits. {\it Static Selection} uses static bit extraction based on value ranges. {\it Greedy Selection} selects 4-bit channels according to error estimation scores. {\it Evolutionary Selection} applies our evolutionary selection algorithm, while {\it Dynamic Extract} dynamically adjusts the bit extraction position. Finally, {\it Finetuning} further trains the models for better mixed-precision accuracy.

Table\,\ref{tab:ablation} presents the accuracy of models under these setups with 75\% 4-bit and 25\% 8-bit mixed-precision quantization. It shows that our bit-lowering is effective even with randomly selected 4-bit channels. Channel prioritization ({\it Greedy}) and evolutionary selection ({\it Evolution}) improve accuracy by an average of 3.5\% and 5.2\%, respectively. Runtime adjustment of bit extraction positions ({\it Dynamic Extract}) adds a 1.1\% improvement, and finetuning gives an additional 0.65\% gain.

\subsection{Case-study: Applying FlexiQ to LLMs}
While FlexiQ is primarily designed for computer vision tasks, we also evaluate its applicability to large language models (LLMs). We first examined unused bits in Qwen and OPT models under 8-bit quantization. Most weight parameters show similar unused-bit patterns to those observed in ViT-Small (Figure\,\ref{fig:unused-bits}), though the last four transformer layers show substantially larger unused bits, and activations across all layers generally have 3–4 unused bits. These characteristics make the models well-suited for FlexiQ.

We then applied quantization to Qwen2.5-0.5B and OPT-350m and measured perplexity on the WikiText2 dataset\,\cite{merity2016pointer}. For OPT-350m, INT8 quantization yielded a perplexity of 27.6, compared to 22.01 for full-precision (bf16). FlexiQ achieved 28.68 (25\% 4-bit), 30.10 (50\%), 33.00 (75\%), and 39.59 (100\% 4-bit), while uniform INT4 quantization degraded sharply to 10938.0. Qwen2.5-0.5B showed similar trends, except that its uniform INT8 perplexity was about nine times higher than the full-precision score. Although techniques such as SmoothQuant\,\cite{XiaoLSWDH23} could improve performance, combining them with FlexiQ is left for future work.

\begin{table}[t]
\centering
\small
\setlength\tabcolsep{3pt}
\renewcommand{\arraystretch}{0.85} 
\caption{Accuracy impact of individual optimizations with 75\% 4-bit and 25\% 8-bit mixed-precision quantization.\label{tab:ablation}}
\begin{tabular}{lrrrr}
\toprule
\textbf{Optimization} & \textbf{RNet18} & \textbf{RNet50} & \textbf{ViT-S} & \textbf{Swin-S} \\ 
\midrule
\textit{Random} & 51.81\% & 66.20\% & 4.06\% & 18.91\% \\ 
\midrule
\textit{+Static Selection} & 57.42\% & 68.30\% & 73.89\% & 80.71\% \\
\textit{+Greedy Selection} & 65.07\% & 72.23\% & 75.37\% & 81.51\% \\
\textit{+Evolutionary Selection} & 68.02\% & 74.69\% & 76.65\% & 81.84\% \\ 
\midrule
\textit{+Dynamic Extract} & 68.81\% & 75.37\% & 78.79\% & 82.72\% \\ 
\midrule
\textit{+Finetuning} & 69.43\% & 76.24\% & 79.54\% & 83.07\% \\
\bottomrule
\end{tabular}
\end{table}

\section{Conclusion}
\label{sec:conclusion}

We present FlexiQ, an adaptive mixed-precision quantization scheme for computer vision models. By selectively applying low-bitwidth computation to feature channels with smaller value ranges and effectively lowering their bitwidth, FlexiQ reduces quantization error while maintaining accuracy on par with high-bitwidth computation.
For efficient execution on NPUs and GPUs, FlexiQ groups multiple feature channels based on the hardware’s processing granularity and assigns their computations accordingly. Moreover, FlexiQ can dynamically adjust its low-bitwidth channel ratio, allowing the quantized models balance accuracy and latency in real-time to handle fluctuating inference workloads.

We implemented FlexiQ prototype, including the mixed-precision inference runtime on NPUs and GPUs. We evaluated FlexiQ using eleven convolution- and transformer-based vision models, focusing on its accuracy and latency trade-offs. It shows that FlexiQ 
achieves on average 6.6\% higher accuracy for 4-bit models with finetuning and outperforms four state-of-the-art quantization techniques. Moreover, our mixed-precision models (25–75\% 4-bit computation) gives an efficient trade-off between inference accuracy and computational load, with the 50\% 4-bit model incurring only 0.6\% accuracy loss on average compared to full-precision models. Finally, latency evaluations on NPUs and GPUs confirmed that FlexiQ incurs minimal runtime overhead, underscoring its hardware efficiency and overall performance benefits.

\section*{Acknowledgement}
This work was supported by 
Institute of Information \& communications Technology Planning \& Evaluation (IITP) grant funded by the Korea government (MSIT) 
(No.RS-2025-02214497 {\footnotesize(25\%)},
No.RS-2025-02263167 {\footnotesize(20\%)},
IITP-2025-II211817 (ITRC) {\footnotesize(18\%)},
No.RS-2024-00438729 {\footnotesize(15\%)},
RS-2021-II211343 {\footnotesize(12\%)},
No.RS-2023-00277060 {\footnotesize(10\%)}). 
This work was also supported by the New Faculty Startup Fund from Seoul National University, the research fund of Hanyang University (HY-201700000002388), and Automation and System Research Institute at Seoul National University (No.0418-20250030).
Jiwon Seo is the corresponding author.

\bibliographystyle{plain}
\bibliography{references}

\begin{thebibliography}{10}

\bibitem{cai2020zeroq}
Yaohui Cai, Zhewei Yao, Zhen Dong, Amir Gholami, Michael~W Mahoney, and Kurt
  Keutzer.
\newblock Zeroq: A novel zero shot quantization framework.
\newblock In {\em Proceedings of the IEEE/CVF Conference on Computer Vision and
  Pattern Recognition}, pages 13169--13178, 2020.

\bibitem{chauhan2023post}
Arun Chauhan, Utsav Tiwari, et~al.
\newblock Post training mixed precision quantization of neural networks using
  first-order information.
\newblock In {\em Proceedings of the IEEE/CVF International Conference on
  Computer Vision}, pages 1343--1352, 2023.

\bibitem{chen2015compressing}
Wenlin Chen, James Wilson, Stephen Tyree, Kilian Weinberger, and Yixin Chen.
\newblock Compressing neural networks with the hashing trick.
\newblock In {\em International conference on machine learning}, pages
  2285--2294. PMLR, 2015.

\bibitem{chmiel2020robust}
Brian Chmiel, Ron Banner, Gil Shomron, Yury Nahshan, Alex Bronstein, Uri
  Weiser, et~al.
\newblock Robust quantization: One model to rule them all.
\newblock {\em Advances in neural information processing systems},
  33:5308--5317, 2020.

\bibitem{ImageNet}
Jia Deng, Wei Dong, Richard Socher, Li-Jia Li, Kai Li, and Li~Fei-Fei.
\newblock Imagenet: A large-scale hierarchical image database.
\newblock In {\em 2009 IEEE Conference on Computer Vision and Pattern
  Recognition}, pages 248--255, 2009.

\bibitem{dettmers2022gpt3}
Tim Dettmers, Mike Lewis, Younes Belkada, and Luke Zettlemoyer.
\newblock Gpt3. int8 (): 8-bit matrix multiplication for transformers at scale.
\newblock {\em Advances in neural information processing systems},
  35:30318--30332, 2022.

\bibitem{dong2020hawq}
Zhen Dong, Zhewei Yao, Daiyaan Arfeen, Amir Gholami, Michael~W Mahoney, and
  Kurt Keutzer.
\newblock Hawq-v2: Hessian aware trace-weighted quantization of neural
  networks.
\newblock {\em Advances in neural information processing systems},
  33:18518--18529, 2020.

\bibitem{dong2019hawq}
Zhen Dong, Zhewei Yao, Amir Gholami, Michael~W Mahoney, and Kurt Keutzer.
\newblock Hawq: Hessian aware quantization of neural networks with
  mixed-precision.
\newblock In {\em Proceedings of the IEEE/CVF International Conference on
  Computer Vision}, pages 293--302, 2019.

\bibitem{fang2021you}
Yuxin Fang, Bencheng Liao, Xinggang Wang, Jiemin Fang, Jiyang Qi, Rui Wu,
  Jianwei Niu, and Wenyu Liu.
\newblock You only look at one sequence: Rethinking transformer in vision
  through object detection.
\newblock {\em Advances in Neural Information Processing Systems},
  34:26183--26197, 2021.

\bibitem{conf/iclr/FrankleC19}
Jonathan Frankle and Michael Carbin.
\newblock The lottery ticket hypothesis: Finding sparse, trainable neural
  networks.
\newblock In {\em 7th International Conference on Learning Representations,
  {ICLR} 2019, New Orleans, LA, USA, May 6-9, 2019}. OpenReview.net, 2019.

\bibitem{ghodrati2020planaria}
Soroush Ghodrati, Byung~Hoon Ahn, Joon~Kyung Kim, Sean Kinzer, Brahmendra~Reddy
  Yatham, Navateja Alla, Hardik Sharma, Mohammad Alian, Eiman Ebrahimi,
  Nam~Sung Kim, et~al.
\newblock Planaria: Dynamic architecture fission for spatial multi-tenant
  acceleration of deep neural networks.
\newblock In {\em 2020 53rd Annual IEEE/ACM International Symposium on
  Microarchitecture (MICRO)}, pages 681--697. IEEE, 2020.

\bibitem{gholami2022survey}
Amir Gholami, Sehoon Kim, Zhen Dong, Zhewei Yao, Michael~W Mahoney, and Kurt
  Keutzer.
\newblock A survey of quantization methods for efficient neural network
  inference.
\newblock In {\em Low-Power Computer Vision}, pages 291--326. Chapman and
  Hall/CRC, 2022.

\bibitem{habi2020hmq}
Hai~Victor Habi, Roy~H Jennings, and Arnon Netzer.
\newblock Hmq: Hardware friendly mixed precision quantization block for cnns.
\newblock In {\em Computer Vision--ECCV 2020: 16th European Conference,
  Glasgow, UK, August 23--28, 2020, Proceedings, Part XXVI 16}, pages 448--463.
  Springer, 2020.

\bibitem{han2015deep}
Song Han, Huizi Mao, and William~J Dally.
\newblock Deep compression: Compressing deep neural networks with pruning,
  trained quantization and huffman coding.
\newblock {\em arXiv preprint arXiv:1510.00149}, 2015.

\bibitem{he2016residual}
Kaiming He, Xiangyu Zhang, Shaoqing Ren, and Jian Sun.
\newblock {Deep Residual Learning for Image Recognition}.
\newblock In {\em Proceedings of 2016 IEEE Conference on Computer Vision and
  Pattern Recognition}, CVPR '16, pages 770--778. IEEE, June 2016.

\bibitem{HowardZCKWWAA17}
Andrew~G. Howard, Menglong Zhu, Bo~Chen, Dmitry Kalenichenko, Weijun Wang,
  Tobias Weyand, Marco Andreetto, and Hartwig Adam.
\newblock Mobilenets: Efficient convolutional neural networks for mobile vision
  applications.
\newblock {\em CoRR}, abs/1704.04861, 2017.

\bibitem{hu2021opq}
Peng Hu, Xi~Peng, Hongyuan Zhu, Mohamed M~Sabry Aly, and Jie Lin.
\newblock Opq: Compressing deep neural networks with one-shot
  pruning-quantization.
\newblock In {\em Proceedings of the AAAI conference on artificial
  intelligence}, volume~35, pages 7780--7788, 2021.

\bibitem{conf/eccv/HuangW18}
Zehao Huang and Naiyan Wang.
\newblock Data-driven sparse structure selection for deep neural networks.
\newblock In Vittorio Ferrari, Martial Hebert, Cristian Sminchisescu, and Yair
  Weiss, editors, {\em Computer Vision - {ECCV} 2018 - 15th European
  Conference, Munich, Germany, September 8-14, 2018, Proceedings, Part {XVI}},
  volume 11220 of {\em Lecture Notes in Computer Science}, pages 317--334.
  Springer, 2018.

\bibitem{jacob2018quantization}
Benoit Jacob, Skirmantas Kligys, Bo~Chen, Menglong Zhu, Matthew Tang, Andrew
  Howard, Hartwig Adam, and Dmitry Kalenichenko.
\newblock Quantization and training of neural networks for efficient
  integer-arithmetic-only inference.
\newblock In {\em Proceedings of the IEEE conference on computer vision and
  pattern recognition}, pages 2704--2713, 2018.

\bibitem{jin2024comprehensive}
Renren Jin, Jiangcun Du, Wuwei Huang, Wei Liu, Jian Luan, Bin Wang, and Deyi
  Xiong.
\newblock A comprehensive evaluation of quantization strategies for large
  language models.
\newblock {\em arXiv preprint arXiv:2402.16775}, 2024.

\bibitem{kim2020robust}
Youngseok Kim, Junyeol Lee, Younghoon Kim, and Jiwon Seo.
\newblock Robust quantization of deep neural networks.
\newblock In {\em Proceedings of the 29th International Conference on Compiler
  Construction}, pages 74--84, 2020.

\bibitem{krizhevsky2009learning}
Alex Krizhevsky.
\newblock Learning multiple layers of features from tiny images.
\newblock pages 32--33, 2009.

\bibitem{lee2021dataflow}
Jounghoo Lee, Jinwoo Choi, Jaeyeon Kim, Jinho Lee, and Youngsok Kim.
\newblock Dataflow mirroring: Architectural support for highly efficient
  fine-grained spatial multitasking on systolic-array npus.
\newblock In {\em 2021 58th ACM/IEEE Design Automation Conference (DAC)}, pages
  247--252. IEEE, 2021.

\bibitem{li2021brecq}
Yuhang Li, Ruihao Gong, Xu~Tan, Yang Yang, Peng Hu, Qi~Zhang, Fengwei Yu, Wei
  Wang, and Shi Gu.
\newblock Brecq: Pushing the limit of post-training quantization by block
  reconstruction.
\newblock {\em arXiv preprint arXiv:2102.05426}, 2021.

\bibitem{li2023alpaserve}
Zhuohan Li, Lianmin Zheng, Yinmin Zhong, Vincent Liu, Ying Sheng, Xin Jin,
  Yanping Huang, Zhifeng Chen, Hao Zhang, Joseph~E Gonzalez, et~al.
\newblock $\{$AlpaServe$\}$: Statistical multiplexing with model parallelism
  for deep learning serving.
\newblock In {\em 17th USENIX Symposium on Operating Systems Design and
  Implementation (OSDI 23)}, pages 663--679, 2023.

\bibitem{lin2023bit}
Chen Lin, Bo~Peng, Zheyang Li, Wenming Tan, Ye~Ren, Jun Xiao, and Shiliang Pu.
\newblock Bit-shrinking: Limiting instantaneous sharpness for improving
  post-training quantization.
\newblock In {\em Proceedings of the IEEE/CVF Conference on Computer Vision and
  Pattern Recognition}, pages 16196--16205, 2023.

\bibitem{lin2024awq}
Ji~Lin, Jiaming Tang, Haotian Tang, Shang Yang, Wei-Ming Chen, Wei-Chen Wang,
  Guangxuan Xiao, Xingyu Dang, Chuang Gan, and Song Han.
\newblock Awq: Activation-aware weight quantization for on-device llm
  compression and acceleration.
\newblock {\em Proceedings of Machine Learning and Systems}, 6:87--100, 2024.

\bibitem{liu2021post}
Xingchao Liu, Mao Ye, Dengyong Zhou, and Qiang Liu.
\newblock Post-training quantization with multiple points: Mixed precision
  without mixed precision.
\newblock In {\em Proceedings of the AAAI conference on artificial
  intelligence}, volume~35, pages 8697--8705, 2021.

\bibitem{Liu_2021_ICCV}
Ze~Liu, Yutong Lin, Yue Cao, Han Hu, Yixuan Wei, Zheng Zhang, Stephen Lin, and
  Baining Guo.
\newblock Swin transformer: Hierarchical vision transformer using shifted
  windows.
\newblock In {\em Proceedings of the IEEE/CVF International Conference on
  Computer Vision (ICCV)}, pages 10012--10022, October 2021.

\bibitem{torchvision2016}
TorchVision maintainers and contributors.
\newblock Torchvision: Pytorch's computer vision library.
\newblock \url{https://github.com/pytorch/vision}, 2016.

\bibitem{merity2016pointer}
Stephen Merity, Caiming Xiong, James Bradbury, and Richard Socher.
\newblock Pointer sentinel mixture models.
\newblock {\em arXiv preprint arXiv:1609.07843}, 2016.

\bibitem{azurepublicdataset}
{Microsoft Azure Public Dataset Contributors}.
\newblock Azure public dataset.
\newblock \url{https://github.com/Azure/AzurePublicDataset}, 2024.

\bibitem{nagel2021white}
Markus Nagel, Marios Fournarakis, Rana~Ali Amjad, Yelysei Bondarenko, Mart
  Van~Baalen, and Tijmen Blankevoort.
\newblock A white paper on neural network quantization.
\newblock {\em arXiv preprint arXiv:2106.08295}, 2021.

\bibitem{googlesla}
NVIDIA.
\newblock Parallel thread execution isa version 8.8.
\newblock \url{https://docs.nvidia.com/cuda/parallel-thread-execution/}.

\bibitem{oh2024exegpt}
Hyungjun Oh, Kihong Kim, Jaemin Kim, Sungkyun Kim, Junyeol Lee, Du-seong Chang,
  and Jiwon Seo.
\newblock Exegpt: Constraint-aware resource scheduling for llm inference.
\newblock In {\em Proceedings of the 29th ACM International Conference on
  Architectural Support for Programming Languages and Operating Systems, Volume
  2}, pages 369--384, 2024.

\bibitem{oh2022out}
Hyungjun Oh, Junyeol Lee, Hyeongju Kim, and Jiwon Seo.
\newblock Out-of-order backprop: An effective scheduling technique for deep
  learning.
\newblock In {\em Proceedings of the Seventeenth European Conference on
  Computer Systems}, pages 435--452, 2022.

\bibitem{peebles2023scalable}
William Peebles and Saining Xie.
\newblock Scalable diffusion models with transformers.
\newblock In {\em Proceedings of the IEEE/CVF international conference on
  computer vision}, pages 4195--4205, 2023.

\bibitem{qu2020adaptive}
Zhongnan Qu, Zimu Zhou, Yun Cheng, and Lothar Thiele.
\newblock Adaptive loss-aware quantization for multi-bit networks.
\newblock In {\em Proceedings of the IEEE/CVF Conference on Computer Vision and
  Pattern Recognition}, pages 7988--7997, 2020.

\bibitem{romero2021infaas}
Francisco Romero, Qian Li, Neeraja~J Yadwadkar, and Christos Kozyrakis.
\newblock $\{$INFaaS$\}$: Automated model-less inference serving.
\newblock In {\em 2021 USENIX Annual Technical Conference (USENIX ATC 21)},
  pages 397--411, 2021.

\bibitem{sharma2016dnnweaver}
Hardik Sharma, Jongse Park, Balavinayagam Samynathan, Behnam Robatmili,
  Shahrzad Mirkhani, and Hadi Esmaeilzadeh.
\newblock Dnnweaver v2. 0: From tensors to fpgas.
\newblock {\em Memory2. 1 ()}, page~3, 2016.

\bibitem{shoeybi2019megatron}
Mohammad Shoeybi, Mostofa Patwary, Raul Puri, Patrick LeGresley, Jared Casper,
  and Bryan Catanzaro.
\newblock Megatron-lm: Training multi-billion parameter language models using
  model parallelism.
\newblock {\em arXiv preprint arXiv:1909.08053}, 2019.

\bibitem{stojkovic2025dynamollm}
Jovan Stojkovic, Chaojie Zhang, {\'I}{\~n}igo Goiri, Josep Torrellas, and Esha
  Choukse.
\newblock Dynamollm: Designing llm inference clusters for performance and
  energy efficiency.
\newblock In {\em 2025 IEEE International Symposium on High Performance
  Computer Architecture (HPCA)}, pages 1348--1362. IEEE, 2025.

\bibitem{thakkar_cutlass_2023}
Vijay Thakkar, Pradeep Ramani, Cris Cecka, Aniket Shivam, Honghao Lu, Ethan
  Yan, Jack Kosaian, Mark Hoemmen, Haicheng Wu, Andrew Kerr, Matt Nicely, Duane
  Merrill, Dustyn Blasig, Fengqi Qiao, Piotr Majcher, Paul Springer, Markus
  Hohnerbach, Jin Wang, and Manish Gupta.
\newblock {CUTLASS}, January 2023.

\bibitem{touvron21a}
Hugo Touvron, Matthieu Cord, Matthijs Douze, Francisco Massa, Alexandre
  Sablayrolles, and Herve Jegou.
\newblock Training data-efficient image transformers \& distillation through
  attention.
\newblock In {\em International Conference on Machine Learning}, volume 139,
  pages 10347--10357, July 2021.

\bibitem{wang2019haq}
Kuan Wang, Zhijian Liu, Yujun Lin, Ji~Lin, and Song Han.
\newblock Haq: Hardware-aware automated quantization with mixed precision.
\newblock In {\em Proceedings of the IEEE/CVF conference on computer vision and
  pattern recognition}, pages 8612--8620, 2019.

\bibitem{pmlr-v119-wang20c}
Peisong Wang, Qiang Chen, Xiangyu He, and Jian Cheng.
\newblock Towards accurate post-training network quantization via bit-split and
  stitching.
\newblock In Hal~Daumé III and Aarti Singh, editors, {\em Proceedings of the
  37th International Conference on Machine Learning}, volume 119 of {\em
  Proceedings of Machine Learning Research}, pages 9847--9856. PMLR, 13--18 Jul
  2020.

\bibitem{wang2023diffusion}
Tangjun Wang, Zehao Dou, Chenglong Bao, and Zuoqiang Shi.
\newblock Diffusion mechanism in residual neural network: theory and
  applications.
\newblock {\em IEEE Transactions on Pattern Analysis and Machine Intelligence},
  46(2):667--680, 2023.

\bibitem{rwightman2021timm}
Ross Wightman.
\newblock timm: Pytorch image models.
\newblock \url{https://github.com/huggingface/pytorch-image-models}, 2021.
\newblock GitHub repository of Hugging Face’s timm library.

\bibitem{DBLP:conf/cvpr/WuLWHC16}
Jiaxiang Wu, Cong Leng, Yuhang Wang, Qinghao Hu, and Jian Cheng.
\newblock Quantized convolutional neural networks for mobile devices.
\newblock In {\em 2016 {IEEE} Conference on Computer Vision and Pattern
  Recognition, {CVPR} 2016, Las Vegas, NV, USA, June 27-30, 2016}, pages
  4820--4828. {IEEE} Computer Society, 2016.

\bibitem{XiaoLSWDH23}
Guangxuan Xiao, Ji~Lin, Micka{\"{e}}l Seznec, Hao Wu, Julien Demouth, and Song
  Han.
\newblock Smoothquant: Accurate and efficient post-training quantization for
  large language models.
\newblock In Andreas Krause, Emma Brunskill, Kyunghyun Cho, Barbara Engelhardt,
  Sivan Sabato, and Jonathan Scarlett, editors, {\em International Conference
  on Machine Learning, {ICML} 2023, 23-29 July 2023, Honolulu, Hawaii, {USA}},
  volume 202 of {\em Proceedings of Machine Learning Research}, pages
  38087--38099. {PMLR}, 2023.

\bibitem{xu2024ptmq}
Ke~Xu, Zhongcheng Li, Shanshan Wang, and Xingyi Zhang.
\newblock Ptmq: Post-training multi-bit quantization of neural networks.
\newblock In {\em Proceedings of the AAAI Conference on Artificial
  Intelligence}, volume~38, pages 16193--16201, 2024.

\bibitem{yang2021fracbits}
Linjie Yang and Qing Jin.
\newblock Fracbits: Mixed precision quantization via fractional bit-widths.
\newblock In {\em Proceedings of the AAAI Conference on Artificial
  Intelligence}, volume~35, pages 10612--10620, 2021.

\bibitem{yao2021hawq}
Zhewei Yao, Zhen Dong, Zhangcheng Zheng, Amir Gholami, Jiali Yu, Eric Tan,
  Leyuan Wang, Qijing Huang, Yida Wang, Michael Mahoney, et~al.
\newblock Hawq-v3: Dyadic neural network quantization.
\newblock In {\em International Conference on Machine Learning}, pages
  11875--11886. PMLR, 2021.

\bibitem{yu2021any}
Haichao Yu, Haoxiang Li, Humphrey Shi, Thomas~S Huang, and Gang Hua.
\newblock Any-precision deep neural networks.
\newblock In {\em Proceedings of the AAAI Conference on Artificial
  Intelligence}, volume~35, pages 10763--10771, 2021.

\bibitem{234998}
Chengliang Zhang, Minchen Yu, Wei Wang, and Feng Yan.
\newblock {MArk}: Exploiting cloud services for {Cost-Effective}, {SLO-Aware}
  machine learning inference serving.
\newblock In {\em 2019 USENIX Annual Technical Conference (USENIX ATC 19)},
  pages 1049--1062, Renton, WA, July 2019. USENIX Association.

\bibitem{zhao2024atom}
Yilong Zhao, Chien-Yu Lin, Kan Zhu, Zihao Ye, Lequn Chen, Size Zheng, Luis
  Ceze, Arvind Krishnamurthy, Tianqi Chen, and Baris Kasikci.
\newblock Atom: Low-bit quantization for efficient and accurate llm serving.
\newblock {\em Proceedings of Machine Learning and Systems}, 6:196--209, 2024.

\bibitem{zheng2021rethinking}
Sixiao Zheng, Jiachen Lu, Hengshuang Zhao, Xiatian Zhu, Zekun Luo, Yabiao Wang,
  Yanwei Fu, Jianfeng Feng, Tao Xiang, Philip~HS Torr, et~al.
\newblock Rethinking semantic segmentation from a sequence-to-sequence
  perspective with transformers.
\newblock In {\em Proceedings of the IEEE/CVF conference on computer vision and
  pattern recognition}, pages 6881--6890, 2021.

\bibitem{zhong2024distserve}
Yinmin Zhong, Shengyu Liu, Junda Chen, Jianbo Hu, Yibo Zhu, Xuanzhe Liu, Xin
  Jin, and Hao Zhang.
\newblock $\{$DistServe$\}$: Disaggregating prefill and decoding for
  goodput-optimized large language model serving.
\newblock In {\em 18th USENIX Symposium on Operating Systems Design and
  Implementation (OSDI 24)}, pages 193--210, 2024.

\end{thebibliography}

\end{document}